\pdfoutput=1

\documentclass[11pt]{article}

\usepackage[table]{xcolor}
\usepackage{acl}

\usepackage{acl}
\usepackage{float}
\usepackage{hyperref}
\usepackage{amsmath} 
\usepackage{amssymb} 
\usepackage{cleveref}
\usepackage{graphicx} 
\usepackage{natbib} 
\usepackage{booktabs} 
\usepackage{array}    %

\usepackage{times}
\usepackage{latexsym}
\usepackage{booktabs}
 \usepackage{amsmath} 
\usepackage[skins]{tcolorbox} 
\usepackage{enumitem}    
\usepackage{geometry}  

\usepackage{multirow}
\usepackage{amssymb}
\usepackage{pifont}
\usepackage{xspace}
\newcommand*{\yoruba}{Yor\`ub\'a\xspace}

\definecolor{bblue}{HTML}{4F81BD}
\definecolor{rred}{HTML}{C0504D}
\definecolor{ggreen}{HTML}{9BBB59}
\usepackage{times}
\usepackage{latexsym}

\usepackage[T1]{fontenc}

\usepackage[utf8]{inputenc}

\usepackage{microtype}

\usepackage{inconsolata}

\usepackage{graphicx}
\usepackage{cleveref}
\usepackage[utf8]{inputenc}
\usepackage{csquotes}

\usepackage{times}
\usepackage{latexsym}
\usepackage{booktabs}
\usepackage{microtype}
\usepackage{amssymb}
\usepackage{pifont}
\usepackage{graphicx}
\usepackage{placeins}
\usepackage{natbib}
\usepackage{longtable}
\usepackage{cleveref}
\usepackage{enumitem}
\usepackage{booktabs}
\usepackage{pdfpages}

\usepackage{tabularx}
\usepackage{multirow}
\usepackage{xspace}

\usepackage{linguex}
\alignSubExtrue

\usepackage[verbose]{newunicodechar}

\newcolumntype{T}{>{\tiny}l}

\usepackage{rotating}
\usepackage{times}
\usepackage{latexsym}
\usepackage{microtype}
\usepackage{amssymb}
\usepackage{pifont}
\usepackage{graphicx}
\usepackage{placeins}
\usepackage{natbib}
\usepackage{tablefootnote}
\usepackage{longtable}
\usepackage{cleveref}
\usepackage{enumitem}
\usepackage{booktabs}
\usepackage{pdfpages}
\usepackage[T1]{fontenc}

\usepackage{tabularx}
\usepackage{multirow}
\usepackage{xspace}
\usepackage{times}
\usepackage{latexsym}
\usepackage[T1]{fontenc}
\usepackage[utf8]{inputenc}
\usepackage{microtype}
\usepackage{inconsolata}
\usepackage{graphicx}
\usepackage{booktabs}
\usepackage{longtable}
\usepackage{lscape}
\usepackage{array}

\usepackage[textsize=footnotesize]{todonotes}
\usepackage{linguex}
\usepackage{subcaption}
\alignSubExtrue
\usepackage[verbose]{newunicodechar}

\usepackage{inconsolata}

\usepackage{makecell}
\global

\tcbset{
  aibox/.style={
    width=474.18663pt,
    top=10pt,
    colback=white,
    colframe=black,
    colbacktitle=black,
    enhanced,
    center,
    attach boxed title to top left={yshift=-0.1in,xshift=0.15in},
    boxed title style={boxrule=0pt,colframe=white,},
  }
}
\newtcolorbox{AIbox}[2][]{aibox,title=#2,#1}

\definecolor{blue1}{RGB}{234, 230, 255} 
\definecolor{blue2}{RGB}{194, 200, 255} 
\definecolor{blue3}{RGB}{154, 170, 255}

\definecolor{red1}{RGB}{255,245,238}
\definecolor{red2}{RGB}{255,228,225}
\definecolor{red3}{RGB}{255,188,185}

\definecolor{light-gray}{HTML}{E5E4E2}
\definecolor{light-cyan}{HTML}{E0FFFF}
\newcolumntype{R}{>{\raggedleft\arraybackslash}p{0.28cm}}
\newcolumntype{L}{>{\raggedleft\arraybackslash}p{0.23cm}}
\newcommand{\extrasmall}{\fontsize{8pt}{10pt}\selectfont}
\newcommand{\extraextrasmall}{\fontsize{7pt}{10pt}\selectfont}

\usepackage{microtype}

\title{SemEval-2025 Task 11: Bridging the Gap in Text-Based Emotion Detection}

\author{Shamsuddeen Hassan Muhammad$^{1,2}$\thanks{Equal contribution}, Nedjma Ousidhoum$^{3*}$, \\
\textbf{Idris Abdulmumin}$^{4}$, \textbf{Seid Muhie Yimam}$^{5}$, \textbf{Jan Philip Wahle}$^{6}$, \textbf{Terry Ruas}$^{6}$, \textbf{Meriem Beloucif}$^{7}$, \\
\textbf{Christine De Kock}$^{8}$, \textbf{Tadesse Destaw Belay}$^{9,10}$, \textbf{Ibrahim Said Ahmad$^{11}$},\textbf{ Nirmal Surange$^{12}$}, \\
\textbf{Daniela Teodorescu$^{13}$}, \textbf{David Ifeoluwa Adelani$^{14,15,16}$}, \textbf{Alham Fikri Aji$^{17}$}, \textbf{Felermino Ali$^{18}$}, \\
\textbf{Vladimir Araujo$^{19}$}, \textbf{Abinew Ali Ayele$^{5,20}$},
\textbf{Oana Ignat$^{21}$}, \textbf{Alexander Panchenko$^{22,23}$}, \textbf{Yi Zhou$^{3}$}, \\
\textbf{Saif M. Mohammad$^{24}$}\\
\footnotesize{$^{1}$Imperial College London, $^{2}$Bayero University Kano, $^3$Cardiff University, $^{4}$DSFSI, 
 University of Pretoria,}\\
 \footnotesize{$^{5}$University of Hamburg, $^{6}$University of Göttingen, $^{7}$Uppsala University, $^{8}$University of Melbourne, $^{9}$Instituto Politécnico Nacional,}\\
 \footnotesize{$^{10}$Wollo University,  $^{11}$Northeastern University,$^{12}$IIIT Hyderabad, $^{13}$University of Alberta, $^{14}$MILA, $^{15}$McGill University,}\\
\footnotesize{$^{16}$Canada CIFAR AI Chair,$^{17}$MBZUAI,$^{18}$LIACC, FEUP, University of Porto, $^{19}$Sailplane AI, $^{20}$Bahir Dar University,}\\
\footnotesize{$^{21}$Santa Clara University, $^{22}$Skoltech, $^{23}$AIRI, $^{24}$National Research Council Canada}\\
       \footnotesize{\texttt{Contact: s.muhammad@imperial.ac.uk, OusidhoumN@cardiff.ac.uk}
 }
 }

\begin{document}
\maketitle
\begin{abstract}

We present our shared task on text-based emotion detection, covering more than 30 languages from seven distinct language families. These languages are predominantly low-resource and are spoken across various continents. The data instances are multi-labeled with six emotional classes, with additional datasets in 11 languages annotated for emotion intensity. Participants were asked to predict labels in three tracks: (a) multilabel emotion detection, (b) emotion intensity score detection, and (c) cross-lingual emotion detection.

The task attracted over 700 participants. We received final submissions from more than 200 teams and 93 system description papers. We report baseline results, along with findings on the best-performing systems, the most common approaches, and the most effective methods across different tracks and languages. The datasets for this task are publicly available.\footnote{The dataset is available at \href{https://brighter-dataset.github.io}{SemEval2025 Task 11}.}

\end{abstract}

\section{Introduction}
People use language in diverse and sophisticated ways to express emotions across languages and cultures \cite{wiebe2005annotating,mohammad-kiritchenko-2018-understanding,mohammad2018semeval}. Emotions are also perceived subjectively, even within the same culture or social group.
Recognising these emotions is central to language technologies and NLP applications in healthcare, digital humanities, dialogue systems, and beyond \cite{mohammad-etal-2018-semeval,saffar2023textual}. In this work, we use \textit{emotion recognition} to refer to \textit{perceived} emotions, i.e., the emotion most people believe the speaker might have felt based on a sentence or short text snippet.
\begin{figure*}[h]
    \centering
     \scalebox{0.88}{
        \includegraphics[width=\linewidth]{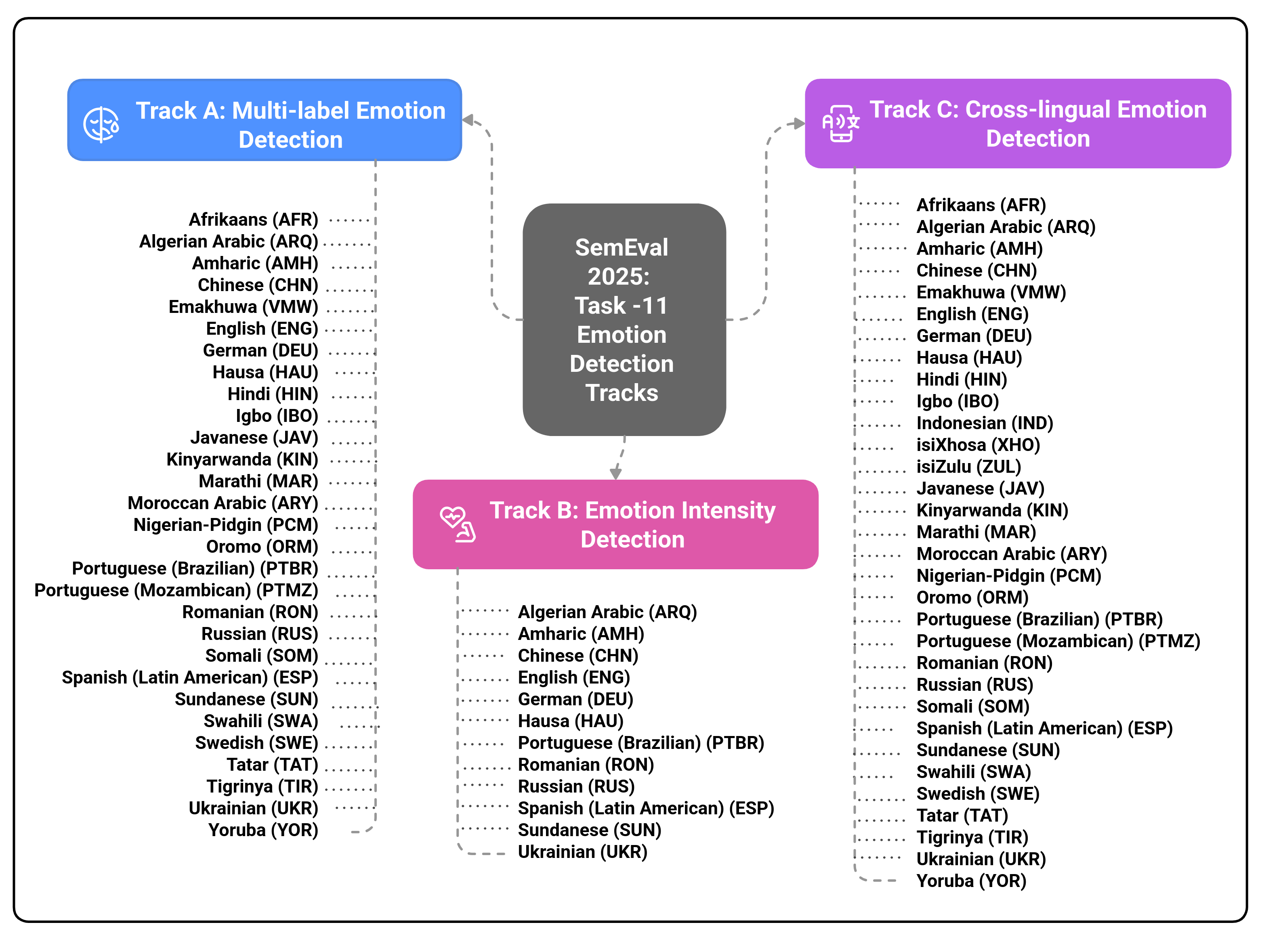}
    }
        \caption{Languages in the three tracks (A, B, and C) of SemEval 2025 Task 11.} 
    
        \label{fig:lgge_fams}
\end{figure*}

Despite the linguistic diversity of regions such as Africa and Asia, which together account for more than 4,000 languages\footnote{\url{https://www.ethnologue.com/insights/how-many-languages}}, few emotion recognition resources exist for these languages. Prior SemEval shared tasks on emotion recognition have primarily focused on high-resource languages such as English, Spanish, and Arabic \cite{strapparava-mihalcea-2007-semeval,mohammad2018semeval,chatterjee-etal-2019-semeval}.
In this task, we provide participants with new datasets covering more than 30 languages from seven distinct language families, spoken across Africa, Asia, Latin America, North America, and Europe \cite{muhammad2025brighterbridginggaphumanannotated}. Our manually annotated emotion recognition datasets, curated in collaboration with local communities, consist of over 100,000 multi-labeled instances drawn from diverse sources, including speeches, social media, news, literature, and reviews. Each instance is labeled by fluent speakers and annotated with six emotion classes: \textit{joy, sadness, anger, fear, surprise, disgust}, and neutral. Additionally, eleven datasets include four emotional intensity levels ranging from 0 to 3 (i.e., absence of emotion to high intensity).

The task consists of three tracks: (a) multilabel emotion detection, (b) emotion intensity detection, and (c) cross-lingual emotion detection. The languages for each track are listed in \Cref{fig:lgge_fams}.
Each team could submit results for one, two, or all three tracks in one or more languages. Our official evaluation metrics were the average F-score for Tracks A and C and the Pearson correlation coefficient for Track B, which measures how well system-predicted intensity scores align with human judgments.

Our task attracted over 700 participants, with 220 final submissions and 93 teams submitting system description papers. Track A (multi-label emotion detection) received the most submissions (114), followed by Track C (cross-lingual emotion detection) with 51, and Track B (emotion intensity detection) with 32. Most teams participated in multiple languages, averaging 11 languages per team.
Our task was the most popular competition on CodaBench in 2024.\footnote{\href{https://github.com/codalab/codabench/wiki/\%5BNewsletter\%5D-CodaLab-in-2024}{Codabench Newsletter 2024}} All task details and resources are available on the task’s GitHub page.\footnote{\url{https://github.com/emotion-analysis-project/SemEval2025-Task11}}
\section{Related Work}
NLP work on emotion detection is predominantly Western-centric, with a few exceptions for languages other than English (e.g., Italian \citep{bianchi-etal-2021-feel}, Romanian \citep{ciobotaru-etal-2022-red}, Indonesian \citep{saputri2018emotion}, and Bengali \citep{iqbal2022bemoc}). While multilingual datasets (e.g., \citep{ohman2020xed} and XLM-EMO \citep{bianchi2022xlm}) exist, they do not fully capture cultural nuances in emotional expressions due to their reliance on translated data (e.g., XLM-EMO), as emotions are highly contextualized and culture-specific \citep{havaldar-etal-2023-multilingual, mohamed-etal-2024-culture, hershcovich-etal-2022-challenges}. Furthermore, most datasets are single-labeled, and to the best of our knowledge, there are no multilingual resources that capture simultaneous emotions and their intensity across various languages.

Additionally, most prior emotion recognition shared tasks have focused on high-resource languages such as English, Spanish, German, and Arabic \cite{strapparava-mihalcea-2007-semeval, mohammad-bravo-marquez-2017-wassa, mohammad-etal-2018-semeval, chatterjee-etal-2019-semeval}. In contrast, this shared task covers more than 30 languages, including several low-resource languages.

\section{Data}
\begin{table}[ht]
\small
\centering
    \resizebox{\linewidth}{!}{
    \begin{tabular}{p{4.1cm}p{0.8cm}p{0.8cm}p{0.8cm}p{0.8cm}}
    \toprule
    \textbf{Language} & \textbf{Train} & \textbf{Dev} & \textbf{Test} & \textbf{Total}\\ 
    \midrule
        Afrikaans (\textbf{\texttt{afr}})  & 2,107 & 98 & 1,065& 3,270\\
        Amharic  (\textbf{\texttt{amh}}) & 3,549 & 592 & 1,774& 5,915 \\
        Algerian Arabic  (\textbf{\texttt{arq}}) & 901 & 100 & 902& 1,903 \\
        Moroccan Arabic  (\textbf{\texttt{ary}})    & 1,608 & 267 & 812& 2,687 \\
        Chinese  (\textbf{\texttt{chn}})   & 2,642 & 200 & 2,642& 5,484 \\
        German  (\textbf{\texttt{deu}})     & 2,603 & 200 & 2,604& 5,407\\
        English  (\textbf{\texttt{eng}}) & 2,768 & 116 & 2,767& 5,651 \\
        Latin American Spanish (\textbf{\texttt{esp}})& 1,996 & 184 & 1,695& 3,875 \\
        Hausa  (\textbf{\texttt{hau}}) & 2,145 & 356 & 1,080&  3,581\\
        Hindi  (\textbf{\texttt{hin}}) & 2,556 & 100 & 1,010& 3,666\\
        Igbo  (\textbf{\texttt{ibo}}) & 2,880 & 479 & 1,444& 4,803 \\
        Indonesian  (\textbf{\texttt{ind}}) & -- & 156 & 851& 1,007 \\
        Javanese  (\textbf{\texttt{jav}}) & -- & 151 & 837 & 988\\
        Kinyarwanda  (\textbf{\texttt{kin}}) &2,451 & 407 & 1,231& 4,089 \\
        Marathi  (\textbf{\texttt{mar}}) & 2,415 & 100 & 1,000& 3,515\\
        Nigerian-Pidgin (\textbf{\texttt{pcm}}) & 3,728 & 620 & 1,870& 6,218 \\
        Oromo  (\textbf{\texttt{orm}}) & 3,442 & 575 & 1,721& 5,738\\
        Portuguese (Brazilian; \textbf{\texttt{ptbr}}) &2,226 & 200 & 2,226& 4,652 \\
        Portuguese (Mozambican; \textbf{\texttt{ptmz}})  & 1,546 & 257 & 776 & 2,579\\
        Romanian (\textbf{\texttt{ron}})   & 1,241
 & 123 & 1,119 & 2,483\\
        Russian (\textbf{\texttt{rus}})   & 2,679 & 199 & 1,000 & 3,878\\
        Somali (\textbf{\texttt{som}})   & 3,392 & 566 & 1,696 & 5,654\\
        Sundanese (\textbf{\texttt{sun}})    & 924 & 199 & 926 & 2,049\\
        Swahili (\textbf{\texttt{swa}})   & 3,307 & 551 & 1,656 & 5,514\\
        Swedish (\textbf{\texttt{swe}})   & 1,187 & 200 & 1,188 & 2,575\\
        Tatar (\textbf{\texttt{tat}})   & 1,000 & 200 & 1,000 & 2,200\\
        Tigrinya (\textbf{\texttt{tir}})   & 3,681 & 614 & 1,840 & 6,135\\
        Ukrainian (\textbf{\texttt{ukr}})   & 2,466 & 249 & 2,234 & 4,949\\
        Emakhuwa (\textbf{\texttt{vmw}})   & 1,551 & 258 & 777 & 2,586\\
        isiXhosa (\textbf{\texttt{xho}})   & -- & 682 & 1,594 & 2,276\\
        Yoruba (\textbf{\texttt{yor}})   & 2,992 & 497 & 1,500 & 4,989\\
        isiZulu (\textbf{\texttt{zul}})   & -- & 875 & 2,047 & 2,922\\
    \bottomrule
    \end{tabular}
    }
    \caption{Languages and data split sizes. Datasets with no training splits (-) were only used in Track C (crosslingual) only.}
    \label{tab:data_sources}
\end{table}

\subsection{Data Collection}
As our task includes more than 30 different datasets, curated and annotated by fluent speakers, we selected data sources based on: 1) the availability of textual data potentially rich in emotions, and 2) access to annotators. Since finding suitable data is challenging when resources are limited, we typically combine sources. The main textual sources used to build our dataset collection are:

\begin{itemize}[noitemsep,nolistsep]
\item \textbf{Social media posts}: Data collected from various platforms, including Reddit (e.g., \texttt{eng}, \texttt{deu}), YouTube (e.g., \texttt{esp}, \texttt{ind}, \texttt{jav}, \texttt{sun}, \texttt{tir}), Twitter (e.g., \texttt{amh}, \texttt{hau}), and Weibo (e.g., \texttt{chn}).

\item \textbf{Personal narratives, talks, speeches}: Anonymised sentences from personal diary posts. We use these in \texttt{eng}, \texttt{deu}, and \texttt{ptbr}, mainly from subreddits such as IAmI. Similarly, the \texttt{afr} dataset includes sentences from speeches and talks.

\item \textbf{Literary texts}: The language lead manually translated the novel \textit{La Grande Maison} (The Big House) by Mohammed Dib\footnote{\url{https://en.wikipedia.org/wiki/La_Grande_Maison}} from French into Algerian Arabic (\texttt{arq}), and post-processed the translation to generate sentences for annotation by native speakers. Note that the translator is bilingual and a native speaker of Algerian Arabic.

\item \textbf{News data}: Although we prefer emotionally rich social media data from different platforms, when such data is scarce, we annotated news data and headlines in some African languages (e.g., \texttt{yor}, \texttt{hau}, and \texttt{vmw}).

\item \textbf{Human-written and machine-generated data}: We created a dataset from scratch for Hindi (\texttt{hin}) and Marathi (\texttt{mar}). Annotators were asked to come up with emotive sentences on a given topic (e.g., family). A small portion of the Hindi dataset was automatically translated into Marathi and manually corrected by native speakers to fix translation errors. Finally, we augmented both datasets with a few hundred quality-approved instances generated by ChatGPT. Note that these constitute less than 1\% of the total number of data instances.

\end{itemize}

\subsection{Data Annotation}
We ask the annotators to select all the emotions that apply to a given text. The set of perceived emotion labels includes: \textit{anger, sadness, fear, disgust, joy, surprise}, and \textit{neutral} (if no emotion is present). The annotators further rate the selected emotion(s) on a four-point intensity scale: 0 (no emotion), 1 (low intensity), 2 (moderate intensity), and 3 (high intensity). We provide the definitions of the categories, annotation guidelines, and more details in \citet{muhammad2025brighterbridginggaphumanannotated}.
We expected some level of disagreement, as emotions are complex, subtle, and perceived differently, even by people within the same culture, especially in the absence of full context. Hence, the final emotion labels were determined based on the emotions and associated intensity values selected by the annotators. Specifically, the given emotion is considered present if:

\begin{enumerate}[noitemsep,nolistsep]
 \item At least two annotators select a label with an intensity value of 1, 2, or 3 (low, medium, or high, respectively). 
\item The average score exceeds a predefined threshold \(T\). We set \(T\) to \(0.5\). 
\end{enumerate}

Once the perceived emotion labels are assigned, the final intensity scores for Track B are determined by averaging the selected intensity values and rounding up to the nearest whole number. Intensity scores are assigned only for datasets in which most instances were annotated by at least five annotators to ensure robustness. \Cref{tab:data_sources} shows the total number of instances in each dataset, as well as the number of instances in the training, development, and test splits for all languages.

\subsection{Annotators' Reliability}
We report the reliability of the annotation using the Split-Half Class Match Percentage (SHCMP; \citealp{mohammad-2024-worrywords}) as described in \citet{muhammad2025brighterbridginggaphumanannotated}. SHCMP extends the concept of Split-Half Reliability (SHR), traditionally used for continuous scores \citep{bws-naacl2016}, to discrete categories like ours (i.e., intensity scores per emotion). Overall, the scores vary from 60\% to more than 90\%, indicating that our datasets are of high quality.

\section{Task Description}
Participants were given text snippets and asked to determine the emotions that people may attribute to the speaker based on a sentence or short text snippet uttered by the speaker. The task consists of three tracks, and participants could participate in one or more of these tracks.

\subsection{Tracks}

\paragraph{Track A: Multi-label Emotion Detection}
Participants were asked to predict the perceived emotion(s) of the speaker and label each text snippet based on the presence (1) or absence (0) of the following emotions: joy, sadness, fear, anger, surprise, and disgust.

\paragraph{Track B: Emotion Intensity Detection}
Given a text and six emotion classes (i.e., joy, sadness, fear, anger, surprise, and disgust), participants were required to predict whether the intensity of each emotion was 0 (no emotion), 1 (low), 2 (medium), or 3 (high). Note that Track B does not include all languages, as intensity scores were only released for datasets with at least five annotators per instance to ensure more robust and reliable labels.

\paragraph{Track C: Cross-lingual Emotion Detection}
Similar to Track A, participants were required to predict the presence or absence of each perceived emotion, but without using any training data in the target language. Instead, they were permitted to use labeled dataset(s) from at least one other language. For instance, one could use German data for training when testing on English. This track focuses on cross-lingual transfer and explores how data from various languages can support emotion detection in low-resource settings, as well as the ability of models to generalise across domains.

\subsection{Task Organisation}
\begin{figure}[!ht]
    \centering
    \includegraphics[trim={0cm 0 0cm 0},clip,width=0.99\linewidth]{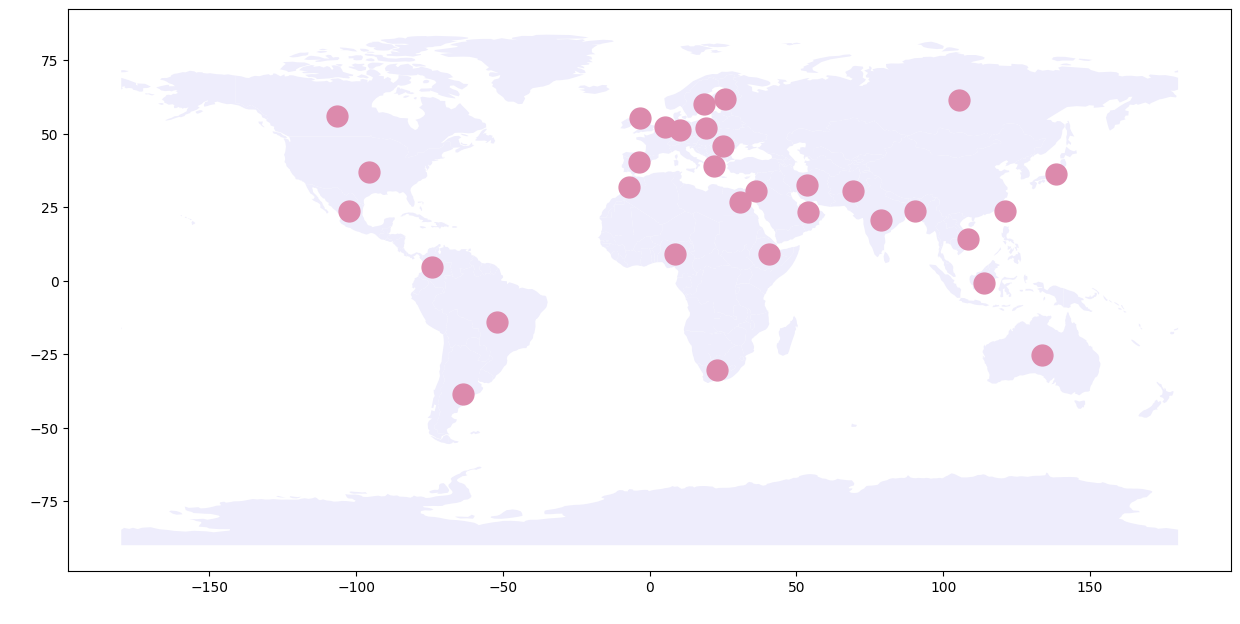}
    \caption{The official affiliations of some of our participants. The list includes 33 countries: Argentina, Australia, Bangladesh, Brazil, Canada, Colombia, Egypt, Ethiopia, Finland, Germany, Greece, India, Indonesia, Iran, Japan, Jordan, Mexico, Morocco, the Netherlands, Nigeria, Pakistan, Poland, Romania, Russia, South Africa, Spain, Sweden, Taiwan, the UAE, the UK, the USA, and Vietnam.}
    \label{fig:participants_map}
\end{figure}
\begin{table*}[ht]
\centering
\footnotesize
\setlength{\tabcolsep}{1.5pt} 
\renewcommand{\arraystretch}{0.75}
\resizebox{\textwidth}{!}{%
\begin{tabular}{llr|llr|llr|llr}
\toprule
\textbf{Lang} & \textbf{Team} & \textbf{Score} & \textbf{Lang} & \textbf{Team} & \textbf{Score} & \textbf{Lang} & \textbf{Team} & \textbf{Score} & \textbf{Lang} & \textbf{Team} & \textbf{Score} \\
\midrule
\textbf{\texttt{afr}} & \cellcolor{blue1}\textbf{\textcolor{black}{pai}} & \cellcolor{blue1}\textbf{0.699} & \textbf{\texttt{amh}} & \cellcolor{blue2}\textbf{\textcolor{black}{chinchunmei}} & \cellcolor{blue2}\textbf{0.773} & \textbf{\texttt{arq}} & \cellcolor{blue1}\textbf{\textcolor{black}{pai}} & \cellcolor{blue1}\textbf{0.669} & \textbf{\texttt{ary}} & \cellcolor{blue1}\textbf{\textcolor{black}{pai}} & \cellcolor{blue1}\textbf{0.629} \\
 & \cellcolor{blue1}\textbf{\textcolor{black}{maomao}} & \cellcolor{blue1}\textbf{0.687} &  & \cellcolor{blue2}\textbf{\textcolor{black}{nust titans}} & \cellcolor{blue2}\textbf{0.714} &  & \cellcolor{blue1}\textbf{\textcolor{black}{jnlp}} & \cellcolor{blue1}\textbf{0.641} &  & \cellcolor{blue1}\textbf{\textcolor{black}{jnlp}} & \cellcolor{blue1}\textbf{0.609} \\
 & \cellcolor{red2}\( R_{\text{baseline}} \) & \cellcolor{red2}0.371 &  & \cellcolor{blue1}\( R_{\text{baseline}} \) & \cellcolor{blue1}0.638 &  & \cellcolor{red1}\( R_{\text{baseline}} \) & \cellcolor{red1}0.414 &  & \cellcolor{red1}\( R_{\text{baseline}} \) & \cellcolor{red1}0.472 \\
 & \cellcolor{red2}\( M_{\text{baseline}}\) & \cellcolor{red2}0.257 &  & \cellcolor{red2}\( M_{\text{baseline}} \) & \cellcolor{red2}0.295 &  & \cellcolor{red1}\( M_{\text{baseline}} \) & \cellcolor{red1}0.445 &  & \cellcolor{red2}\( M_{\text{baseline}} \) & \cellcolor{red2}0.247 \\
\midrule

\textbf{\texttt{chn}} & \cellcolor{blue2}\textbf{\textcolor{black}{pai}} & \cellcolor{blue2}\textbf{0.709} & \textbf{\texttt{deu}} & \cellcolor{blue2}\textbf{\textcolor{black}{pai}} & \cellcolor{blue2}\textbf{0.740} & \textbf{\texttt{eng}} & \cellcolor{blue3}\textbf{\textcolor{black}{pai}} & \cellcolor{blue3}\textbf{0.823} & \textbf{\texttt{esp}} & \cellcolor{blue3}\textbf{\textcolor{black}{pai}} & \cellcolor{blue3}\textbf{0.849} \\
 & \cellcolor{blue1}\textbf{\textcolor{black}{teleai}} & \cellcolor{blue1}\textbf{0.682} &  & \cellcolor{blue2}\textbf{\textcolor{black}{heimerdinger}} & \cellcolor{blue2}\textbf{0.706} &  & \cellcolor{blue3}\textbf{\textcolor{black}{nycu-nlp}} & \cellcolor{blue3}\textbf{0.823} &  & \cellcolor{blue3}\textbf{\textcolor{black}{heimerdinger}} & \cellcolor{blue3}\textbf{0.838} \\
 & \cellcolor{red1}\( R_{\text{baseline}} \) & \cellcolor{red1}0.531 &  & \cellcolor{blue1}\( R_{\text{baseline}} \) & \cellcolor{blue1}0.642 &  & \cellcolor{blue2}\( R_{\text{baseline}} \) & \cellcolor{blue2}0.708 &  & \cellcolor{blue2}\( R_{\text{baseline}} \) & \cellcolor{blue2}0.774 \\
 & \cellcolor{red2}\( M_{\text{baseline}} \) & \cellcolor{red2}0.278 &  & \cellcolor{red1}\( M_{\text{baseline}} \) & \cellcolor{red1}0.449 &  & \cellcolor{red2}\( M_{\text{baseline}} \) & \cellcolor{red2}0.367 &  & \cellcolor{red2}\( M_{\text{baseline}} \) & \cellcolor{red2}0.312 \\ \midrule
 
\textbf{\texttt{hau}} & \cellcolor{blue2}\textbf{\textcolor{black}{pai}} & \cellcolor{blue2}\textbf{0.751} & \textbf{\texttt{hin}} & \cellcolor{blue3}\textbf{\textcolor{black}{jnlp}} & \cellcolor{blue3}\textbf{0.926} & \textbf{\texttt{ibo}} & \cellcolor{blue1}\textbf{\textcolor{black}{pai}} & \cellcolor{blue1}\textbf{0.600} & \textbf{\texttt{kin}} & \cellcolor{blue1}\textbf{\textcolor{black}{pai}} & \cellcolor{blue1}\textbf{0.657} \\
 & \cellcolor{blue1}\textbf{\textcolor{black}{empaths}} & \cellcolor{blue1}\textbf{0.695} &  & \cellcolor{blue3}\textbf{\textcolor{black}{pai}} & \cellcolor{blue3}\textbf{0.920} &  & \cellcolor{red1}\textbf{\textcolor{black}{late-gil-nlp}} & \cellcolor{red1}\textbf{0.563} &  & \cellcolor{red1}\textbf{\textcolor{black}{mcgill-nlp}} & \cellcolor{red1}\textbf{0.590} \\
 & \cellcolor{red1}\( R_{\text{baseline}} \) & \cellcolor{red1}0.596 &  & \cellcolor{blue3}\( R_{\text{baseline}} \) & \cellcolor{blue3}0.855 &  & \cellcolor{red1}\( R_{\text{baseline}} \) & \cellcolor{red1}0.479 &  & \cellcolor{red1}\( R_{\text{baseline}} \) & \cellcolor{red1}0.463 \\
 & \cellcolor{red2}\( M_{\text{baseline}} \) & \cellcolor{red2}0.312 &  & \cellcolor{red2}\( M_{\text{baseline}} \) & \cellcolor{red2}0.246 &  & \cellcolor{red2}\( M_{\text{baseline}} \) & \cellcolor{red2}0.236 &  & \cellcolor{red2}\( M_{\text{baseline}} \) & \cellcolor{red2}0.218 \\ \midrule
 
\textbf{\texttt{mar}} & \cellcolor{blue3}\textbf{\textcolor{black}{pai}} & \cellcolor{blue3}\textbf{0.884} & \textbf{\texttt{orm}} & \cellcolor{blue1}\textbf{\textcolor{black}{tewodros}} & \cellcolor{blue1}\textbf{0.616} & \textbf{\texttt{pcm}} & \cellcolor{blue1}\textbf{\textcolor{black}{pai}} & \cellcolor{blue1}\textbf{0.674} & \textbf{\texttt{ptbr}} & \cellcolor{blue1}\textbf{\textcolor{black}{pai}} & \cellcolor{blue1}\textbf{0.683} \\
 & \cellcolor{blue3}\textbf{\textcolor{black}{indidataminer}} & \cellcolor{blue3}\textbf{0.883} & & \cellcolor{red1}\textbf{\textcolor{black}{late-gil-nlp}} & \cellcolor{red1}\textbf{0.592} & & \cellcolor{blue1}\textbf{\textcolor{black}{jnlp}} & \cellcolor{blue1}\textbf{0.634} & & \cellcolor{blue1}\textbf{\textcolor{black}{heimerdinger}} & \cellcolor{blue1}\textbf{0.625} \\
 & \cellcolor{blue3}\( R_{\text{baseline}} \) & \cellcolor{blue3}0.822 &  & \cellcolor{red3}\( R_{\text{baseline}} \) & \cellcolor{red3}0.126 &  & \cellcolor{red1}\( R_{\text{baseline}} \) & \cellcolor{red1}0.555 &  & \cellcolor{red1}\( R_{\text{baseline}} \) & \cellcolor{red1}0.426 \\ 
 & \cellcolor{red2}\( M_{\text{baseline}} \) & \cellcolor{red2}0.264 &  & \cellcolor{red2}\( M_{\text{baseline}} \) & \cellcolor{red2}0.232 &  & \cellcolor{red2}\( M_{\text{baseline}} \) & \cellcolor{red2}0.357 &  & \cellcolor{red2}\( M_{\text{baseline}} \) & \cellcolor{red2}0.243 \\ 
 \midrule
 
\textbf{\texttt{ptmz}} & \cellcolor{red1}\textbf{\textcolor{black}{pai}} & \cellcolor{red1}\textbf{0.548} & \textbf{\texttt{ron}} & \cellcolor{blue2}\textbf{\textcolor{black}{pai}} & \cellcolor{blue2}\textbf{0.794} & \textbf{\texttt{rus}} & \cellcolor{blue3}\textbf{\textcolor{black}{heimerdinger}} & \cellcolor{blue3}\textbf{0.901} & \textbf{\texttt{som}} & \cellcolor{red1}\textbf{\textcolor{black}{pai}} & \cellcolor{red1}\textbf{0.577} \\
 & \cellcolor{red1}\textbf{\textcolor{black}{heimerdinger}} & \cellcolor{red1}\textbf{0.507} & & \cellcolor{blue2}\textbf{\textcolor{black}{jnlp}} & \cellcolor{blue2}\textbf{0.779} & & \cellcolor{blue3}\textbf{\textcolor{black}{jnlp}} & \cellcolor{blue3}\textbf{0.891} & & \cellcolor{red1}\textbf{\textcolor{black}{empaths}} & \cellcolor{red1}\textbf{0.508} \\
 & \cellcolor{red1}\( R_{\text{baseline}} \) & \cellcolor{red1}0.459 &  & \cellcolor{blue2}\( R_{\text{baseline}} \) & \cellcolor{blue2}0.762 &  & \cellcolor{blue3}\( R_{\text{baseline}} \) & \cellcolor{blue3}0.838 &  & \cellcolor{red1}\( R_{\text{baseline}} \) & \cellcolor{red1}0.459 \\ 
 & \cellcolor{red3}\( M_{\text{baseline}} \) & \cellcolor{red3}0.163 &  & \cellcolor{red1}\( M_{\text{baseline}} \) & \cellcolor{red1}0.461 &  & \cellcolor{red2}\( M_{\text{baseline}} \) & \cellcolor{red2}0.262 &  & \cellcolor{red3}\( M_{\text{baseline}} \) & \cellcolor{red3}0.198 \\
\midrule
\textbf{\texttt{sun}} & \cellcolor{red1}\textbf{\textcolor{black}{lazarus nlp}} & \cellcolor{red1}\textbf{0.550} & \textbf{\texttt{swa}} & \cellcolor{red2}\textbf{\textcolor{black}{empaths}} & \cellcolor{red2}\textbf{0.386} & \textbf{\texttt{swe}} & \cellcolor{blue1}\textbf{\textcolor{black}{pai}} & \cellcolor{blue1}\textbf{0.626} & \textbf{\texttt{tat}} & \cellcolor{blue3}\textbf{\textcolor{black}{pai}} & \cellcolor{blue3}\textbf{0.846} \\
 & \cellcolor{red1}\textbf{\textcolor{black}{pai}} & \cellcolor{red1}\textbf{0.541} & & \cellcolor{red2}\textbf{\textcolor{black}{pai}} & \cellcolor{red2}\textbf{0.385} & & \cellcolor{blue1}\textbf{\textcolor{black}{jnlp}} & \cellcolor{blue1}\textbf{0.619} & & \cellcolor{blue2}\textbf{\textcolor{black}{tue-jms}} & \cellcolor{blue2}\textbf{0.797} \\
 & \cellcolor{red2}\( R_{\text{baseline}} \) & \cellcolor{red2}0.373 &  & \cellcolor{red2}\( R_{\text{baseline}} \) & \cellcolor{red2}0.227 &  & \cellcolor{red1}\( R_{\text{baseline}} \) & \cellcolor{red1}0.520 &  & \cellcolor{red1}\( R_{\text{baseline}} \) & \cellcolor{red1}0.539 \\ 
 & \cellcolor{red2}\( M_{\text{baseline}} \) & \cellcolor{red2}0.334 &  & \cellcolor{red3}\( M_{\text{baseline}} \) & \cellcolor{red3}0.179 &  & \cellcolor{red2}\( M_{\text{baseline}} \) & \cellcolor{red2}0.264 &  & \cellcolor{red2}\( M_{\text{baseline}} \) & \cellcolor{red2}0.246 \\

\midrule
\textbf{\texttt{tir}} & \cellcolor{red1}\textbf{\textcolor{black}{nta}} & \cellcolor{red1}\textbf{0.591} & \textbf{\texttt{ukr}} & \cellcolor{blue2}\textbf{\textcolor{black}{pai}} & \cellcolor{blue2}\textbf{0.726} & \textbf{\texttt{vmw}} & \cellcolor{red2}\textbf{\textcolor{black}{team unibuc}} & \cellcolor{red2}\textbf{0.325} & \textbf{\texttt{yor}} & \cellcolor{red1}\textbf{\textcolor{black}{pai}} & \cellcolor{red1}\textbf{0.461} \\
 & \cellcolor{red1}\textbf{\textcolor{black}{late-gil-nlp}} & \cellcolor{red1}\textbf{0.587} &  & \cellcolor{blue1}\textbf{\textcolor{black}{csiro-lt}} & \cellcolor{blue1}\textbf{0.664} &  & \cellcolor{red2}\textbf{\textcolor{black}{pai}} & \cellcolor{red2}\textbf{0.255} &  & \cellcolor{red2}\textbf{\textcolor{black}{heimerdinger}} & \cellcolor{red2}\textbf{0.392} \\
 & \cellcolor{red1}$ R_{\text{baseline}} $ & \cellcolor{red1}0.463 &  & \cellcolor{red1}$ R_{\text{baseline}} $ & \cellcolor{red1}0.535 &  & \cellcolor{red3}$ R_{\text{baseline}} $ & \cellcolor{red3}0.121 &  & \cellcolor{red3}$ R_{\text{baseline}} $ & \cellcolor{red3}0.092 \\
 & \cellcolor{red2}$ M_{\text{baseline}} $ & \cellcolor{red2}0.253 &  & \cellcolor{red3}$ M_{\text{baseline}} $ & \cellcolor{red3}0.157 &  & \cellcolor{red3}$ M_{\text{baseline}} $ & \cellcolor{red3}0.163 &  & \cellcolor{red3}$ M_{\text{baseline}} $ & \cellcolor{red3}0.165 \\

 \bottomrule
\end{tabular}%
}
\caption{Average macro-F1 scores for our baselines ($M_{baseline}$ and $R_{baseline}$, referring to the Majority Vote and RoBERTa baselines, respectively) and the top two performing systems in Track A (shown in bold) for each language.
}
\label{tab:top2teams_trackA}
\end{table*}

We used Codabench as the competition platform and released pilot datasets before the start of the shared task to help participants better understand the task (i.e., the datasets, the languages involved, and the labels). We provided participants with a starter kit on GitHub, resources for beginners, and organised a Q\&A session along with a writing tutorial for junior researchers. Our participants were based in different parts of the world, as shown in \Cref{fig:participants_map}, with many coming from underrepresented regions.
The task consisted of two phases: (1) the development phase and (2) the evaluation phase. During the development phase, the leaderboard was open, allowing a maximum of 999 submissions per participant. In the evaluation phase, the leaderboard was closed, and each participant was allowed up to three submissions, with the last submission being considered for the official ranking.

\subsection{Evaluation Metrics and Baselines}
\paragraph{Evaluation Metrics}
For Tracks A and C, we use the average macro F-score calculated based on the predicted and the gold-standard labels. For Track B, we use the Pearson correlation coefficient, which captures how well the system-predicted intensity scores of test instances align with human judgments. We provided the participants with an evaluation script on our GitHub page.
\paragraph{Our Baselines}
We run a simple majority class baseline for each language across all three tracks. Further, for Tracks A and B (Tables \ref{tab:top2teams_trackA} and \ref{ab:top2teams_trackB}, respectively), we fine-tuned RoBERTa using the training data for each language. \Cref{tab:top2teams_trackA} shows the average macro F-scores of the top-performing systems compared to our baseline in Track A, and \Cref{ab:top2teams_trackB} shows the Pearson correlation scores for Track B.
For Track C (\Cref{tab:top2teams_trackC}), we fine-tuned RoBERTa by training on all languages within a language family while holding out one target language used for testing, e.g., all Indo-European languages except \texttt{eng} when testing on it. For language families with only one language, we trained on the Slavic languages (\texttt{rus} and \texttt{ukr}) and tested on \texttt{tat}; on the Niger-Congo languages (\texttt{swa} and \texttt{yor}) and tested on \texttt{pcm}; and trained on \texttt{rus} when testing on \texttt{chn}.

\section{Participating Systems and Results}

\subsection{Overview}
Our task attracted more than 700 registered participants and was featured in the Codabench newsletter as the most popular competition hosted on Codabench in 2024. 

In the development phase, 153 submissions were made for Track A, 52 for Track B, and 25 for Track C. In the test phase, 220 submissions were made for Track A, 96 for Track B, and 46 for Track C. The official results include more than 220 final submissions from 93 teams. 
While the English subtracks received the highest number of submissions, we note that other languages, including underserved ones, were comparable in terms of popularity. 

We report results only for teams that submitted a system description paper. \Cref{tab:track_a} presents the results for Track A, which had 87 participating teams. \Cref{tab:track_b} shows the results for Track B, with 38 participating teams, while \Cref{tab:track_c} reports the results for Track C, which had 21 participating teams.

\subsection{Track A: Multi-label Emotion Detection}

\subsubsection{Best-Performing Systems}

\paragraph{Team Pai} proposes one of the most effective models in the competition. They consistently rank as the top approach in Track A for 20 out of 28 languages. For their system, they combine several base models (ChatGPT-4o \cite{openai2024gpt4o}, DeepSeek-V3 \cite{deepseekai2025deepseekr1incentivizingreasoningcapability}, Gemma-9b \cite{team2024gemma}, Qwen-2.5-32b \cite{yang2024qwen2}, Mistral-Small-24B \cite{jiang2024mixtral}) using multiple ensemble techniques (neural networks, XGBoost, LightGBM, linear regression, weighted voting). They fine-tune Gemma-9b and Qwen-2.5-32b using AdaLoRA. For prompting the LLMs, they used an iterative prompt-optimisation technique that generates prompt variations.

\paragraph{Team Chinchunmei} ranks in the top 10 in 16 languages in Track A and 12th in English. They use sample contrastive learning, where performance is enhanced by comparing sample pairs, and generative contrastive learning, where the models learn to distinguish correct from incorrect predictions. Their samples are randomly selected from the task dataset (no external augmentation). They use LLaMa3-Instruct-8B \cite{llama3modelcard} for their fine-tuning.

\subsubsection{Takeaways}
Most of the teams that rank well on Track A experiment with essentially two methodologies: 1) fine-tuning BERT-based models such as DeBERTa \cite{siino-2024-deberta}, mBERT \cite{dolev-2023-mbert}, and XLM-R \cite{conneau-etal-2020-unsupervised}; and/or 2) instruction-fine-tuning using LoRa methodologies in combination with prompt design and data augmentation techniques on LLMs (ChatGPT-4o, DeepSeek-V3, Gemma-9b, Qwen-2.5-32b, Mistral-Small-24B). For instance, Team Telai (3rd in Chinese), Team Empaths (2nd in Hausa and Somali, and 1st in Swahili), Team NYCU-NLP (2nd in English), Team JNLP (1st in Hindi and among the best four across 10 languages), Team Unibouc (1st in Emakhuwa), Team Heiderdinger (2nd in Mozambican Portuguese, German, Spanish, Brazilian Portuguese, and \yoruba; 1st in Russian), and Team Maomao (2nd in Afrikaans).

Few teams focus on only a subset of languages or explore language-related knowledge in their methodology. For instance, Team Lazarus NLP redefined and reformulated the multi-label classification into multiple binary tasks to expand training samples. They also explored how knowledge could potentially be transferred between Indonesian languages.
All the top 10 teams performed significantly better than our baseline model, with an improvement that is more notable in a few low-resource languages, such as Oromo, where Team Tewodros obtained an average macro-F1 score of 0.616 compared to a baseline of 0.126. The same was observed for \yoruba, where Team Pai scored 0.461 compared to a baseline score of 0.092.

\subsection{Track B:  Emotion Intensity Detection}
\begin{table}[]
\centering
\resizebox{\linewidth}{!}{
\renewcommand{\arraystretch}{0.9}  
\setlength{\tabcolsep}{3pt}
\begin{tabular}{llr|llr}
\toprule
\textbf{Lang} & \textbf{Team} & \textbf{Score} & \textbf{Lang} & \textbf{Team} & \textbf{Score} \\
\midrule
\textbf{\texttt{amh}} & \cellcolor{blue2}\textbf{csecu-learners} & \cellcolor{blue2}\textbf{0.856} & \textbf{\texttt{arq}} & \cellcolor{blue1}\textbf{pai} & \cellcolor{blue1}\textbf{0.650} \\
& \cellcolor{blue2}\textbf{heimerdinger} & \cellcolor{blue2}\textbf{0.781} &  & \cellcolor{red1}\textbf{jnlp} & \cellcolor{red1}\textbf{0.587} \\

 & \cellcolor{red1}\( R_{\text{baseline}} \) & \cellcolor{red1}0.508 &  & \cellcolor{red3}$R_{\text{baseline}}$ & \cellcolor{red3}0.016 \\

& \cellcolor{red3}$M_{\text{baseline}}$ & \cellcolor{red3}-0.001 &  & \cellcolor{red3}\( M_{\text{baseline}} \) & \cellcolor{red3}-0.009 \\ \midrule

\textbf{\texttt{chn}} & \cellcolor{blue2}\textbf{pai} & \cellcolor{blue2}\textbf{0.722} & \textbf{\texttt{deu}} & \cellcolor{blue2}\textbf{pai} & \cellcolor{blue2}\textbf{0.766} \\
& \cellcolor{blue2}\textbf{teleai} & \cellcolor{blue2}\textbf{0.708} &  & \cellcolor{blue2}\textbf{teleai} & \cellcolor{blue2}\textbf{0.743} \\
& \cellcolor{red1}$R_{\text{baseline}}$ & \cellcolor{red1}0.405 &  &  \cellcolor{red1}$R_{\text{baseline}}$& \cellcolor{red1}0.562 \\ 

 & \cellcolor{red3}$M_{\text{baseline}}$ & \cellcolor{red3}0.000 &  & \cellcolor{red3}$M_{\text{baseline}}$ & \cellcolor{red3}0.016 \\ \midrule

\textbf{\texttt{eng}} & \cellcolor{blue3}\textbf{pai} & \cellcolor{blue3}\textbf{0.840} & \textbf{\texttt{esp}} & \cellcolor{blue3}\textbf{pai} & \cellcolor{blue3}\textbf{0.808} \\
& \cellcolor{blue3}\textbf{nycu-nlp} & \cellcolor{blue3}\textbf{0.837} &  & \cellcolor{blue2}\textbf{deepwave} & \cellcolor{blue2}\textbf{0.792} \\
& \cellcolor{blue1}$R_{\text{baseline}}$ & \cellcolor{blue1}0.641 &  & \cellcolor{blue2}$R_{\text{baseline}}$ & \cellcolor{blue2}0.726 \\ 

 & \cellcolor{red3}$M_{\text{baseline}}$ & \cellcolor{red3}0.001 &  & \cellcolor{red3}$M_{\text{baseline}}$ & \cellcolor{red3}0.011 \\ \midrule

\textbf{\texttt{hau}} & \cellcolor{blue2}\textbf{pai} & \cellcolor{blue2}\textbf{0.770} & \textbf{\texttt{ptbr}} & \cellcolor{blue2}\textbf{pai} & \cellcolor{blue2}\textbf{0.710} \\
& \cellcolor{blue2}\textbf{deepwave} & \cellcolor{blue2}\textbf{0.747} &  & \cellcolor{blue2}\textbf{teleai} & \cellcolor{blue2}\textbf{0.690} \\
& \cellcolor{red2}$R_{\text{baseline}}$ & \cellcolor{red2}0.270 &  & \cellcolor{red2}$R_{\text{baseline}}$ & \cellcolor{red2}0.297 \\ 

& \cellcolor{red3}$M_{\text{baseline}}$ & \cellcolor{red3}0.003 &  & \cellcolor{red3}$M_{\text{baseline}}$ & \cellcolor{red3}0.016 \\ \midrule

\textbf{\texttt{ron}} & \cellcolor{blue2}\textbf{pai} & \cellcolor{blue2}\textbf{0.726} & \textbf{\texttt{rus}} & \cellcolor{blue3}\textbf{pai} & \cellcolor{blue3}\textbf{0.925} \\
& \cellcolor{blue2}\textbf{deepwave} & \cellcolor{blue2}\textbf{0.716} &  & \cellcolor{blue3}\textbf{teleai} & \cellcolor{blue3}\textbf{0.919} \\
& \cellcolor{red1}$R_{\text{baseline}}$ & \cellcolor{red1}0.557 &  & \cellcolor{blue3}$R_{\text{baseline}}$ & \cellcolor{blue3}0.877 \\ 

& \cellcolor{red3}$M_{\text{baseline}}$ & \cellcolor{red3}0.003 &  & \cellcolor{red3}$M_{\text{baseline}}$ & \cellcolor{red3}0.016 \\ \midrule

\textbf{\texttt{ukr}} & \cellcolor{blue2}\textbf{pai} & \cellcolor{blue2}\textbf{0.708} & \multicolumn{3}{c}{\multirow{3}{*}{}} \\
& \cellcolor{blue2}\textbf{jnlp} & \cellcolor{blue2}\textbf{0.672} \\
& \cellcolor{red1}$R_{\text{baseline}}$ & \cellcolor{red1}0.399 \\

& \cellcolor{red3}$M_{\text{baseline}}$ & \cellcolor{red3}-0.01  \\ 
\bottomrule
\end{tabular}
}
\caption{Pearson correlation scores for our baselines (Majority: $M_{baseline}$ and RoBERTa: $R_{baseline}$) and the top two performing systems in Track B (shown in bold) for each language.
}
\label{ab:top2teams_trackB}
\end{table}

\begin{table*}[ht]
\centering
\footnotesize
\setlength{\tabcolsep}{1.5pt} 
\renewcommand{\arraystretch}{0.75} 
\resizebox{\textwidth}{!}{%
\begin{tabular}{@{}llr|llr|llr|llr@{}}
\toprule
\textbf{Lang} & \textbf{Team} & \textbf{Score} & \textbf{Lang} & \textbf{Team} & \textbf{Score} & \textbf{Lang} & \textbf{Team} & \textbf{Score} & \textbf{Lang} & \textbf{Team} & \textbf{Score} \\
\midrule
\textbf{\texttt{afr}} & \cellcolor{blue2}maomao & \cellcolor{blue2}\textbf{0.705} & \textbf{\texttt{amh}} & \cellcolor{blue1}\textbf{deepwave} & \cellcolor{blue1}\textbf{0.661} & \textbf{\texttt{arq}} & \cellcolor{red1}\textbf{deepwave} & \cellcolor{red1}\textbf{0.588} & \textbf{\texttt{ary}} & \cellcolor{blue1}\textbf{deepwave} & \cellcolor{blue1}\textbf{0.632} \\
 & \cellcolor{red1}\textbf{deepwave} & \cellcolor{red1}\textbf{0.574} & & \cellcolor{blue1}\textbf{uob-nlp} & \cellcolor{blue1}\textbf{0.627} & & \cellcolor{red1}\textbf{maomao} & \cellcolor{red1}\textbf{0.584} & & \cellcolor{red1}\textbf{maomao} & \cellcolor{red1}\textbf{0.565} \\
 & \cellcolor{red2}$R_{\text{baseline}}$ & \cellcolor{red2}0.350 &  & \cellcolor{red1}$R_{\text{baseline}}$ & \cellcolor{red1}0.487 & &  \cellcolor{red2}$R_{\text{baseline}}$ & \cellcolor{red2}0.338 &  & \cellcolor{red2}$R_{\text{baseline}}$ & \cellcolor{red2}0.355 \\ 
 & \cellcolor{red2}$M_{\text{baseline}}$ & \cellcolor{red2}0.257 &  & \cellcolor{red2}$M_{\text{baseline}}$ & \cellcolor{red2}0.295 && \cellcolor{red1}$M_{\text{baseline}}$ & \cellcolor{red1}0.445 &  &\cellcolor{red2}$M_{\text{baseline}}$ & \cellcolor{red2}0.247 \\ \midrule
\textbf{\texttt{chn}} & \cellcolor{blue1}\textbf{deepwave} & \cellcolor{blue1}\textbf{0.689} & \textbf{\texttt{deu}} & \cellcolor{blue2}\textbf{deepwave} & \cellcolor{blue2}\textbf{0.727} & \textbf{\texttt{eng}} & \cellcolor{blue2}\textbf{deepwave} & \cellcolor{blue2}\textbf{0.797} & \textbf{\texttt{esp}} & \cellcolor{blue3}\textbf{deepwave} & \cellcolor{blue3}\textbf{0.831} \\
 & \cellcolor{blue1}\textbf{maomao} & \cellcolor{blue1}\textbf{0.622} & & \cellcolor{blue1}\textbf{gt-nlp} & \cellcolor{blue1}\textbf{0.687} & & \cellcolor{blue2}\textbf{maomao} & \cellcolor{blue2}\textbf{0.755} & & \cellcolor{blue3}\textbf{maomao} & \cellcolor{blue3}\textbf{0.806} \\
 & \cellcolor{red2}$R_{\text{baseline}}$ & \cellcolor{red2}0.246 &  & \cellcolor{red1}$R_{\text{baseline}}$ & \cellcolor{red1}0.468 &  & \cellcolor{red2}$R_{\text{baseline}}$ & \cellcolor{red2}0.375 &  & \cellcolor{red1}$R_{\text{baseline}}$ & \cellcolor{red1}0.574 \\ 
 & \cellcolor{red2}$M_{\text{baseline}}$ & \cellcolor{red2}0.278 &  & \cellcolor{red2}$M_{\text{baseline}}$ & \cellcolor{red2}0.319 && \cellcolor{red1}$M_{\text{baseline}}$ & \cellcolor{red1}0.449 &  & \cellcolor{red2}$M_{\text{baseline}}$ & \cellcolor{red2}0.367 \\ \midrule
\textbf{\texttt{hau}} & \cellcolor{blue2}\textbf{deepwave} & \cellcolor{blue2}\textbf{0.709} & \textbf{\texttt{hin}} & \cellcolor{blue3}\textbf{deepwave} & \cellcolor{blue3}\textbf{0.919} & \textbf{\texttt{ibo}} & \cellcolor{blue1}\textbf{deepwave} & \cellcolor{blue1}\textbf{0.605} & \textbf{\texttt{ind}} & \cellcolor{blue1}\textbf{maomao} & \cellcolor{blue1}\textbf{0.672} \\
 & \cellcolor{blue1}\textbf{uob-nlp} & \cellcolor{blue1}\textbf{0.627} & & \cellcolor{blue3}\textbf{maomao} & \cellcolor{blue3}\textbf{0.896} & & \cellcolor{red1}\textbf{uob-nlp} & \cellcolor{red1}\textbf{0.484} & & \cellcolor{blue1}\textbf{lazarus nlp} & \cellcolor{blue1}\textbf{0.641} \\
 & \cellcolor{red2}$R_{\text{baseline}}$ & \cellcolor{red2}0.320 &  & \cellcolor{red3}$R_{\text{baseline}}$ & \cellcolor{red3}0.138 &  & \cellcolor{red3}$R_{\text{baseline}}$ & \cellcolor{red3}0.075 &  & \cellcolor{red2}$R_{\text{baseline}}$ & \cellcolor{red2}0.376 \\ 
 & \cellcolor{red2}$M_{\text{baseline}}$ & \cellcolor{red2}0.312 &  & \cellcolor{red2}$M_{\text{baseline}}$ & \cellcolor{red2}0.264 && \cellcolor{red2}$M_{\text{baseline}}$ & \cellcolor{red2}0.236 &  & \cellcolor{red3}$M_{\text{baseline}}$ & \cellcolor{red3}0.254 \\ \midrule
\textbf{\texttt{jav}} & \cellcolor{red1}\textbf{heimerdinger} & \cellcolor{red1}\textbf{0.439} & \textbf{\texttt{kin}} & \cellcolor{red1}\textbf{deepwave} & \cellcolor{red1}\textbf{0.508} & \textbf{\texttt{mar}} & \cellcolor{blue3}\textbf{deepwave} & \cellcolor{blue3}\textbf{0.903} & \textbf{\texttt{orm}} & \cellcolor{red1}\textbf{deepwave} & \cellcolor{red1}\textbf{0.542} \\
 & \cellcolor{red1}\textbf{lazarus nlp} & \cellcolor{red1}\textbf{0.438} & & \cellcolor{red1}\textbf{uob-nlp} & \cellcolor{red1}\textbf{0.466} & & \cellcolor{blue3}\textbf{maomao} & \cellcolor{blue3}\textbf{0.863} & & \cellcolor{red1}\textbf{uob-nlp} & \cellcolor{red1}\textbf{0.491} \\
 & \cellcolor{red1}$R_{\text{baseline}}$ & \cellcolor{red1}0.464 &  & \cellcolor{red3}$R_{\text{baseline}}$ & \cellcolor{red3}0.184 &  & \cellcolor{blue2}$R_{\text{baseline}}$ & \cellcolor{blue2}0.772 &  & \cellcolor{red2}$R_{\text{baseline}}$ & \cellcolor{red2}0.262 \\ 
 & \cellcolor{red2}$M_{\text{baseline}}$ & \cellcolor{red2}0.204 &  & \cellcolor{red2}$M_{\text{baseline}}$ & \cellcolor{red2}0.218 && \cellcolor{red2}$M_{\text{baseline}}$ & \cellcolor{red2}0.264 &  & \cellcolor{red2}$M_{\text{baseline}}$ & \cellcolor{red2}0.232 \\ \midrule
\textbf{\texttt{pcm}} & \cellcolor{blue1}\textbf{deepwave} & \cellcolor{blue1}\textbf{0.674} & \textbf{\texttt{ptbr}} & \cellcolor{blue1}\textbf{deepwave} & \cellcolor{blue1}\textbf{0.629} & \textbf{\texttt{ptmz}} & \cellcolor{red1}\textbf{deepwave} & \cellcolor{red1}\textbf{0.555} & \textbf{\texttt{ron}} & \cellcolor{blue2}\textbf{deepwave} & \cellcolor{blue2}\textbf{0.767} \\
 & \cellcolor{red1}\textbf{maomao} & \cellcolor{red1}\textbf{0.562} & & \cellcolor{blue1}\textbf{maomao} & \cellcolor{blue1}\textbf{0.617} & & \cellcolor{red1}\textbf{maomao} & \cellcolor{red1}\textbf{0.495} & & \cellcolor{blue2}\textbf{maomao} & \cellcolor{blue2}\textbf{0.747} \\
 & \cellcolor{red3}$R_{\text{baseline}}$ & \cellcolor{red3}0.010 &  & \cellcolor{red1}$R_{\text{baseline}}$ & \cellcolor{red1}0.418 &  & \cellcolor{red2}$R_{\text{baseline}}$ & \cellcolor{red2}0.297 &  & \cellcolor{blue2}$R_{\text{baseline}}$ & \cellcolor{blue2}0.762 \\ 
 & \cellcolor{red2}$M_{\text{baseline}}$ & \cellcolor{red2}0.357 &  & \cellcolor{red2}$M_{\text{baseline}}$  & \cellcolor{red2}0.243 && \cellcolor{red3}$M_{\text{baseline}}$  & \cellcolor{red3}0.163 &  & \cellcolor{blue2}$M_{\text{baseline}}$  & \cellcolor{blue2}0.652 \\ \midrule
\textbf{\texttt{rus}} & \cellcolor{blue3}\textbf{deepwave} & \cellcolor{blue3}\textbf{0.906} & \textbf{\texttt{som}} & \cellcolor{red1}\textbf{maomao} & \cellcolor{red1}\textbf{0.488} & \textbf{\texttt{sun}} & \cellcolor{red1}\textbf{deepwave} & \cellcolor{red1}\textbf{0.467} & \textbf{\texttt{swa}} & \cellcolor{red2}\textbf{maomao} & \cellcolor{red2}\textbf{0.381} \\
 & \cellcolor{blue3}\textbf{maomao} & \cellcolor{blue3}\textbf{0.852} & & \cellcolor{red1}\textbf{deepwave} & \cellcolor{red1}\textbf{0.488} & & \cellcolor{red1}\textbf{maomao} & \cellcolor{red1}\textbf{0.464} & & \cellcolor{red2}\textbf{deepwave} & \cellcolor{red2}\textbf{0.355} \\
 & \cellcolor{blue2}$R_{\text{baseline}}$  & \cellcolor{blue2}0.704 &  & \cellcolor{red2}$R_{\text{baseline}}$& \cellcolor{red2}0.273 &  & \cellcolor{red3}$R_{\text{baseline}}$ & \cellcolor{red3}0.194 &  & \cellcolor{red3}$R_{\text{baseline}}$ & \cellcolor{red3}0.190 \\ 
 & \cellcolor{red2}$M_{\text{baseline}}$ & \cellcolor{red2}0.262 &  & \cellcolor{red3}$M_{\text{baseline}}$ & \cellcolor{red3}0.198 && \cellcolor{red2}$M_{\text{baseline}}$ & \cellcolor{red2}0.334 &  & \cellcolor{red3}$M_{\text{baseline}}$ & \cellcolor{red3}0.179 \\ \midrule
\textbf{\texttt{swe}} & \cellcolor{blue1}\textbf{deepwave} & \cellcolor{blue1}\textbf{0.645} & \textbf{\texttt{tat}} & \cellcolor{blue2}\textbf{deepwave} & \cellcolor{blue2}\textbf{0.789} & \textbf{\texttt{tir}} & \cellcolor{red1}\textbf{deepwave} & \cellcolor{red1}\textbf{0.505} & \textbf{\texttt{ukr}} & \cellcolor{blue2}\textbf{deepwave} & \cellcolor{blue2}\textbf{0.702} \\
 & \cellcolor{red1}\textbf{maomao} & \cellcolor{red1}\textbf{0.578} & & \cellcolor{blue1}\textbf{maomao} & \cellcolor{blue1}\textbf{0.697} & & \cellcolor{red1}\textbf{uob-nlp} & \cellcolor{red1}\textbf{0.445} & & \cellcolor{blue1}\textbf{maomao} & \cellcolor{blue1}\textbf{0.623} \\
 & \cellcolor{red1}$R_{\text{baseline}}$ & \cellcolor{red1}0.512 &  & \cellcolor{red1}$R_{\text{baseline}}$ & \cellcolor{red1}0.445 &  & \cellcolor{red2}$R_{\text{baseline}}$ & \cellcolor{red2}0.339 &  & \cellcolor{red1}$R_{\text{baseline}}$ & \cellcolor{red1}0.496 \\ 
 & \cellcolor{red2}$M_{\text{baseline}}$ & \cellcolor{red2}0.264 &  & \cellcolor{red2}$M_{\text{baseline}}$ & \cellcolor{red2}0.246 && \cellcolor{red2}$M_{\text{baseline}}$ & \cellcolor{red2}0.253 &  & \cellcolor{red3}$M_{\text{baseline}}$ & \cellcolor{red3}0.157 \\ \midrule
\textbf{\texttt{vmw}} & \cellcolor{red2}\textbf{deepwave} & \cellcolor{red2}\textbf{0.210} & \textbf{\texttt{xho}} & \cellcolor{red1}\textbf{maomao} & \cellcolor{red1}\textbf{0.443} & \textbf{\texttt{yor}} & \cellcolor{red2}\textbf{maomao} & \cellcolor{red2}\textbf{0.360} & \textbf{\texttt{zul}} & \cellcolor{red2}\textbf{maomao} & \cellcolor{red2}\textbf{0.397} \\
 & \cellcolor{red3}\textbf{ozemi} & \cellcolor{red3}\textbf{0.193} & & \cellcolor{red2}\textbf{ozemi} & \cellcolor{red2}\textbf{0.315} & & \cellcolor{red2}\textbf{deepwave} & \cellcolor{red2}\textbf{0.342} & & \cellcolor{red2}\textbf{heimerdinger} & \cellcolor{red2}\textbf{0.226} \\
 & \cellcolor{red3}$R_{\text{baseline}}$ & \cellcolor{red3}0.052 &  & \cellcolor{red3}$R_{\text{baseline}}$ & \cellcolor{red3}0.127 &  & \cellcolor{red3}$R_{\text{baseline}}$ & \cellcolor{red3}0.053 &  & \cellcolor{red3}$R_{\text{baseline}}$ & \cellcolor{red3}0.153 \\ 
 & \cellcolor{red3}$M_{\text{baseline}}$ & \cellcolor{red3}0.162 &  & \cellcolor{red3}$M_{\text{baseline}}$ & \cellcolor{red3}0.115 && \cellcolor{red3}$M_{\text{baseline}}$ & \cellcolor{red3}0.165 &  & \cellcolor{red3}$M_{\text{baseline}}$ & \cellcolor{red3}0.109 \\ 
\bottomrule
\end{tabular}%
}
\caption{Average macro-F1 scores for our baselines ($M_{baseline}$ and $R_{baseline}$, referring to the Majority Vote and RoBERTa baselines, respectively) and the top two performing systems in Track C (shown in bold) for each language.
}
\label{tab:top2teams_trackC}
\end{table*}

\subsubsection{Best-Performing Systems}
\paragraph{Team Pai}
Similar to Track A, Team Pai ranked at the top across all languages in Track B, except for Amharic. They used an ensemble of LLMs, combining several base models (ChatGPT-4o\cite{openai2024gpt4o}, DeepSeek-V3 \cite{deepseekai2025deepseekr1incentivizingreasoningcapability}, Gemma-9b \cite{team2024gemma}, Qwen-2.5-32b \cite{yang2024qwen2}, Mistral-Small-24B) with multiple ensemble techniques (neural networks, XGBoost, LightGBM, linear regression, weighted voting). They fine-tuned the Gemma and Qwen models using AdaLoRA. 
For prompting the LLMs, they employed an iterative prompt-optimisation technique to generate prompt variations.

\paragraph{Team CSECU-Learners}
CSECU-Learners ranked at the top in Amharic by fine-tuning language-specific transformers (XLM-Roberta \cite{conneau-etal-2020-unsupervised} for Amharic) with a classification layer and multi-sample dropout.
\subsubsection{Takeaways}
Teams Deepwave, Teleai, and JNLP also ranked highly across various languages using prompt engineering approaches similar to those in Track A. Additionally, Team NYCU-NLP ranked second in English by aggregating instruction-tuned small language models. All these teams outperformed our RoBERTa baseline, which achieved moderate Pearson correlation coefficient scores overall, but performed poorly in languages such as Algerian Arabic, Hausa, Ukrainian, and even Brazilian Portuguese -highlighting the difficulty of the task.

Overall, we observe that most teams adopted approaches similar to those used in Track A, with only minor adjustments to the prompts. Notably, even the best-performing teams achieved a Pearson correlation coefficient of no more than 0.65 on Algerian Arabic, likely due to the novelty and complexity of the dataset.
\subsection{Track C: Cross-lingual Emotion Detection}
\begin{figure*}[!ht]
    \centering
    \includegraphics[trim={10 0 10 0},clip,width=.8\textwidth]{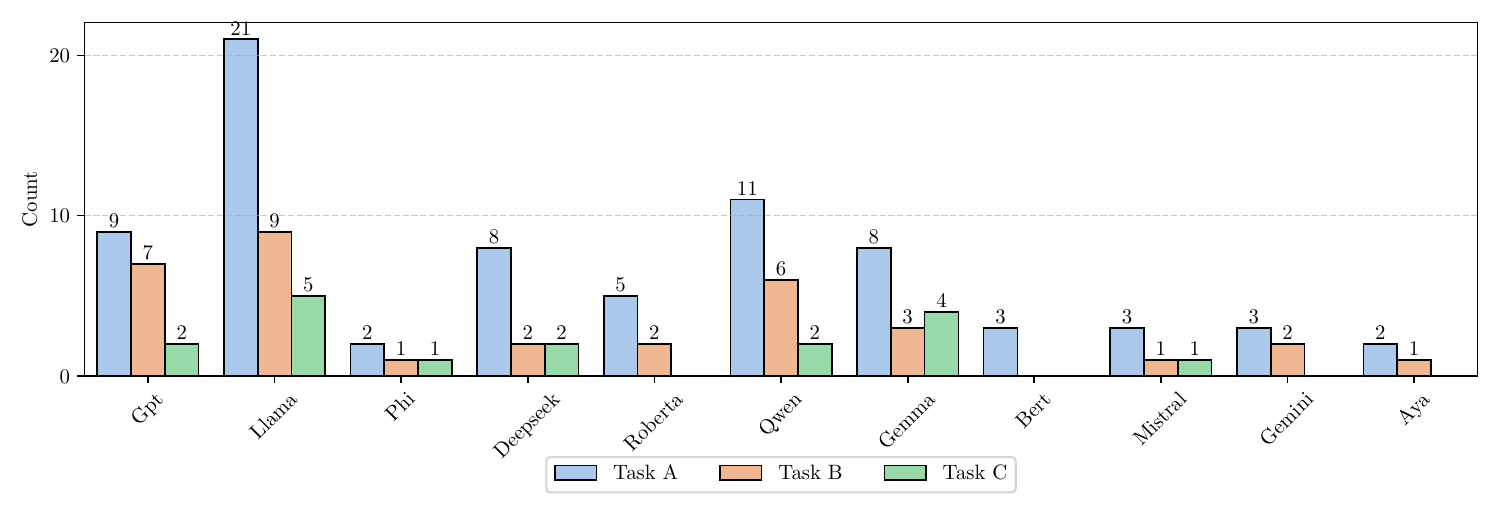}
    \caption{Top LLMs used by participants across tracks (A, B, and C).}
    \label{fig:llm_used}
\end{figure*}
\subsubsection{Best-Performing Systems}

\paragraph{Team deepwave}
Team Deepwave fine-tuned Google Gemma-2 \cite{team2024gemma} using tailored data augmentation and Chain-of-Thought (CoT) prompting. They decomposed the task into two sub-tasks: (1) sentiment keyword identification and (2) sentiment polarity recognition.

To address the challenge of limited data, they employed k-fold (k=5) cross-validation and used model merging—a strategy that combines the predictions of multiple models to improve generalization—by averaging the prediction probabilities of each model, assigning equal weights of 0.2. In this track, a dedicated LoRA module was trained for each target language. The training dataset for each module comprised data from all other languages in Track A, excluding the target language $L_i$. They exclusively used augmented data generated through CoT prompting for training.

\paragraph{Team maomao}
Team maomao experimented with different setups for fine-tuning LLMs. The base models used were Qwen2.5-7B-Instruct, GPT-0544o Mini2, and LLaMA-3.2-3B-Instruct \citep{llama3modelcard}. They applied Direct Preference Optimisation to refine their model -a technique that selects high-quality instances from lower-quality ones within a dataset. After this step, they retrained the model using the refined dataset. They also explored random sampling and retrieval-augmented generation (RAG) methods for training, primarily on DeepSeek-V3, Qwen-Max5093, and Grok-V2.

\subsubsection{Takeaways}

Other top-performing systems include Team Ozemi, who fine-tuned a multilingual BERT model and applied machine translation to enhance performance across all languages. They used the Synthetic Minority Oversampling Technique (SMOTE) with TF-IDF to address class imbalance in Russian. They translated all the datasets into a common language using Google Translate before processing -except for Nigerian Pidgin and Emakhua, where they used a multilingual BERT model for translation. They also leveraged two Kaggle competition datasets for data augmentation.

Team heimerdinger, who ranked highest in Javanese, built their approach using various LLMs (LLaMA 3.1 8B, Qwen 2.5 7B, DeepSeek-7B, MistralV0.3-7B, and Gemma2-9B) for Track C. They employed in-context learning with multilingual examples from high-resource languages such as English and Spanish.

Overall, the participating teams outperformed our baselines. However, the average scores for this track are notably lower, particularly in low-resource languages, due to the additional challenges posed by limited data and resources. As shown in Table \ref{tab:top2teams_trackC}, there are significant performance gaps -even the top systems did not achieve an F-score higher than 0.50 in languages such as Javanese, Somali, Sundanese, Xhosa, \yoruba, isiZulu, and Emakhuwa (where the top system achieved an F-score of only 0.21).

\section{Discussion}

\paragraph{Popular Methods}
Unsurprisingly, most top-performing teams favored fine-tuning and prompting large language models (LLMs) such as Gemma-2, Mistral, Phi-4, Qwen-2.5, DeepSeek, LLaMA-3, GPT, and Gemini models.
For fine-tuning, both full fine-tuning and parameter-efficient fine-tuning were the most commonly used strategies to enhance performance.

For prompting, few-shot, zero-shot, and chain-of-thought prompting were the most frequently used techniques.

Many participants also experimented with traditional transformer-based models, particularly XLM-RoBERTa, mBERT, DeBERTa, and IndicBERT \cite{kakwani2020indicnlpsuite} (see \Cref{fig:llm_used} and \Cref{fig:methods_overview} in the Appendix).

\paragraph{Best Performing Systems}
The results from the top-performing submissions suggest that while LLMs achieve strong overall performance, their effectiveness is heavily dependent on prompt engineering techniques and wording.

Additionally, performance varies significantly by language. Across all tracks, LLM-based approaches and the best-performing systems consistently yielded better results for high-resource languages such as English and Russian. In contrast, performance dropped notably when tested on low-resource languages such as Swahili and Emakhuwa.

Furthermore, most teams did not incorporate additional datasets to enhance performance (see Appendix), as few-shot and zero-shot approaches proved highly effective.

\section{Conclusion}
We presented our shared task on text-based emotion recognition, which covered three tracks and a total of 32 languages. The submitted systems were ranked based on macro F1-scores for Tracks A and C, comparing predicted labels to gold labels, and based on the ranking of predicted intensity scores for Track B.

We summarised the reported results, discussing the best-performing and most innovative methods.

Overall, performance varied significantly across languages. Our results highlight that emotion recognition remains an open challenge, particularly for under-served languages and in low-resource settings.

\section{Limitations} \label{sec:limitations}
Emotions are subjective and subtle, and they are expressed and perceived differently. We do not claim that our datasets capture the true emotions of the speakers, fully represent language use across the 32 languages, or cover all possible emotions. We discuss the ethical considerations extensively in \Cref{sec:ethics}.

We acknowledge the limited data sources available for some low-resource languages. Therefore, our datasets may not be suitable for tasks requiring large amounts of data in a given language. However, they serve as a valuable starting point for research in this area.

\section{Ethical Considerations} \label{sec:ethics}
Emotion perception and expression are subjective and nuanced, as they are influenced by various factors (e.g., cultural background, social group, personal experiences, and social context).
Thus, it is impossible to determine someone's emotions with absolute certainty based solely on short text snippets.
Our datasets explicitly focus on perceived emotions—identifying the emotions that most people believe the speaker may have felt.
We do not claim to annotate the speaker’s true emotions, as these cannot be definitively determined from text alone.
We recognise the importance of this distinction, as perceived emotions may differ from actual emotions.

We acknowledge potential biases in our data, as we rely on text-based communication, which inherently carries biases from data sources and annotators.
Additionally, while many of our datasets focus on low-resource languages, we do not claim they fully represent these languages' usage. Further, although we took measures to filter inappropriate content, some instances may have been overlooked. 

We explicitly urge careful consideration of ethical implications before using our datasets. We prohibit their use for commercial purposes or by state actors in high-risk applications unless explicitly approved by the dataset creators.
Systems built on our datasets may not be reliable at the individual instance level and are susceptible to domain shifts.
Thus, they should not be used for critical decision-making, such as in health applications, without expert supervision.
For a more in-depth discussion, see \citet{mohammad-2022-ethics,mohammad-2023-best}.

Finally, all annotators involved in the study were compensated above the minimum hourly wage. 

\section*{Acknowledgments}
Shamsuddeen Muhammad acknowledges the support of Google DeepMind and Lacuna Fund, an initiative co-founded by The Rockefeller Foundation, Google.org, and Canada’s International Development Research Centre. The views expressed herein do not necessarily represent those of Lacuna Fund, its Steering Committee, its funders, or Meridian Institute.
\newline
Nedjma Ousidhoum would like to thank Abderrahmane Samir Lazouni, Lyes Taher Khalfi, Manel Amarouche, Narimane Zahra Boumezrag, Noufel Bouslama, Abderaouf Ousidhoum, Sarah Arab, Wassil Adel Merzouk, Yanis Rabai, and another annotator who would like to stay anonymous for their work and insightful comments. 
\newline
Idris Abdulmumin gratefully acknowledges the ABSA UP Chair of Data Science for funding his post-doctoral research and providing compute resources.
\newline
Jan Philip Wahle and Terry Ruas were partially supported by the Lower Saxony Ministry of Science and Culture and the
VW Foundation.
\newline
Meriem Beloucif acknowledges Nationella Språkbanken (the Swedish National Language Bank) and Swe-CLARIN – jointly funded by the Swedish Research Council (2018–2024; dnr 2017-00626) and its 10 partner institutions for funding the Swedish annotations.
\newline
Alexander Panchenko would like to thank Nikolay Ivanov, Artem Vazhentsev, Mikhail Salnikov, Maria Marina, Vitaliy Protasov, Sergey Pletenev, Daniil Moskovskiy, Vasiliy Konovalov, Elisey Rykov, and Dmitry Iarosh for their help with the annotation for Russian. Preparation of Tatar data was funded by AIRI and completed by Dina Abdullina, Marat Shaehov, and Ilseyar Alimova.
\newline
Yi Zhou would like to thank Gaifan Zhang, Bing Xiao, and Rui Qin for their help with the annotations and for providing feedback.






\bibliography{anthology,custom,semeval-task-11}

\appendix

\section{Task Results and Analysis}
\begin{figure*}
    \centering
    \begin{subfigure}{\textwidth}
        \centering
        \includegraphics[width=.95\textwidth, clip, trim=10 0 10 0]{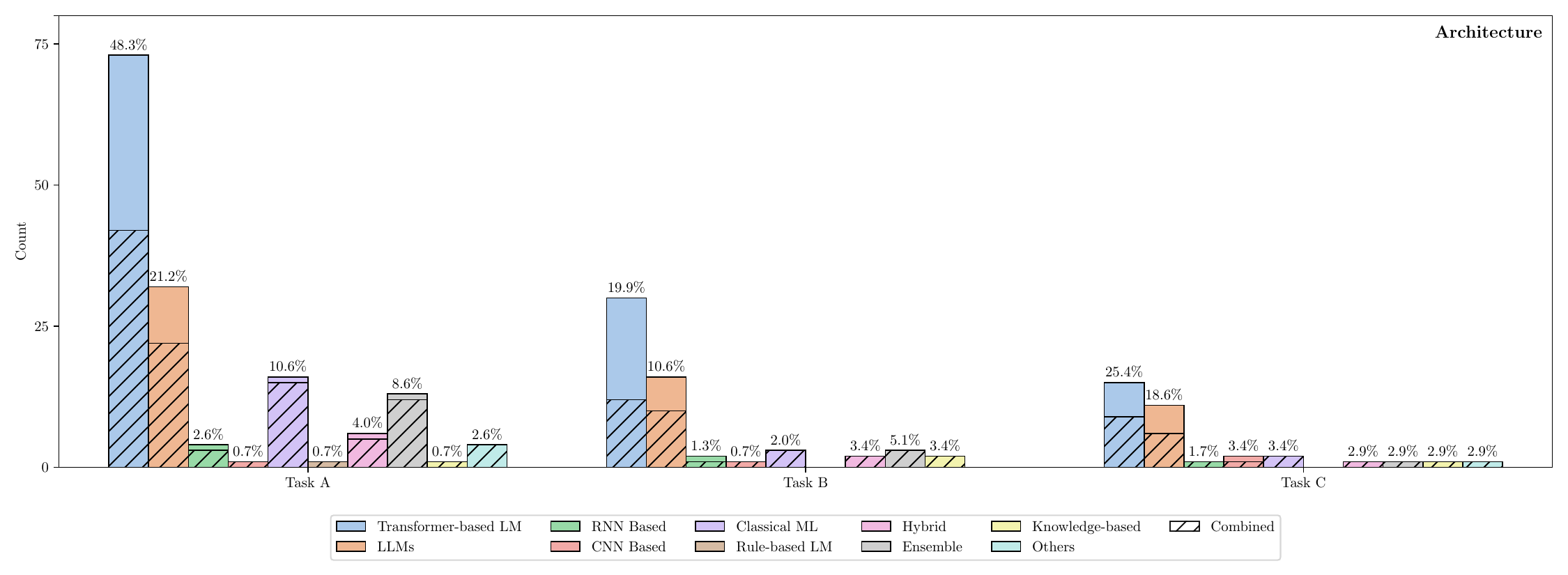}
        \label{fig:popular_architectures}
    \end{subfigure}
    
    \begin{subfigure}{\textwidth}
        \centering
        \includegraphics[width=.95\textwidth, clip, trim=30 0 54 0]{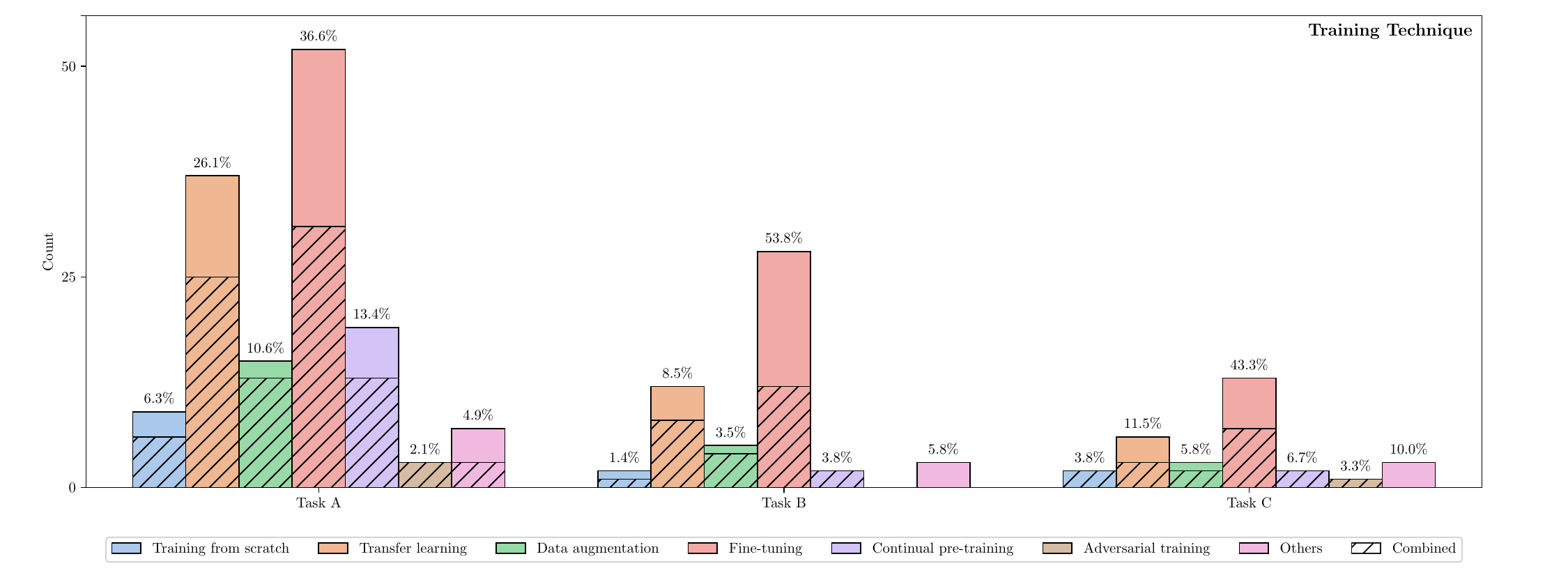}
        \label{fig:training_techniques}
    \end{subfigure}
    
    \begin{subfigure}{\textwidth}
        \centering
        \includegraphics[width=.95\textwidth, clip, trim=10 0 10 0]{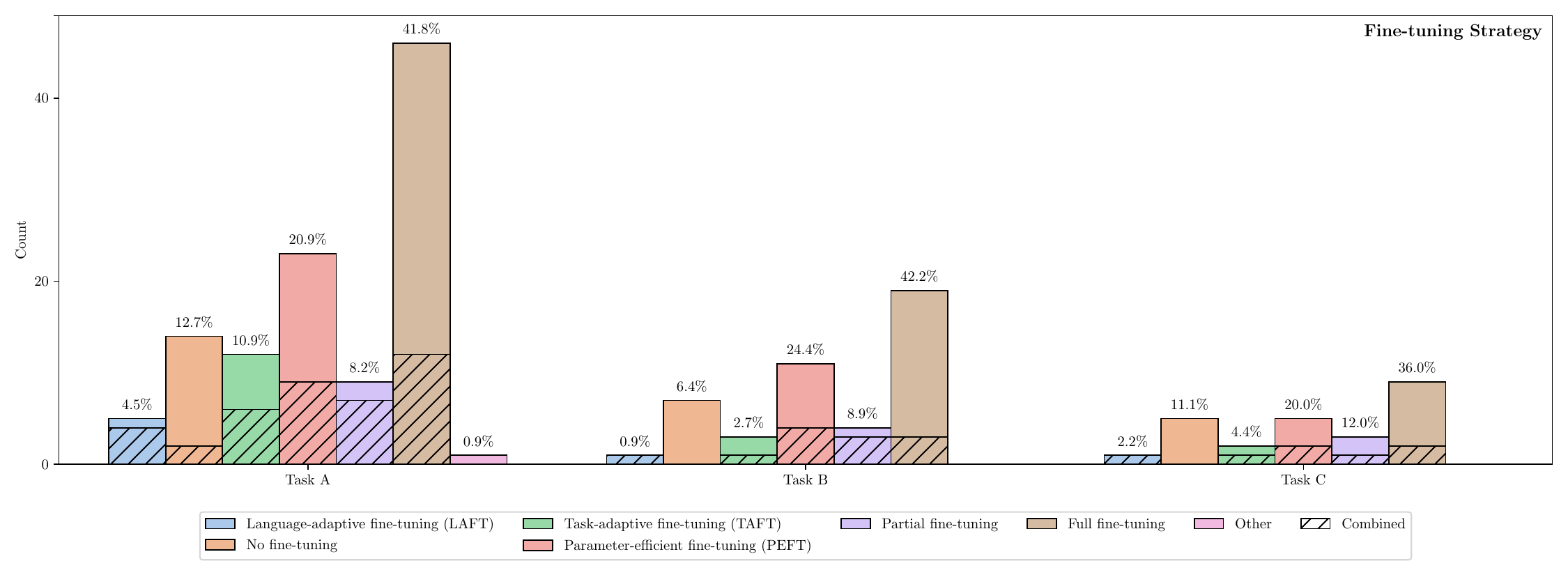}
        \label{fig:finetune_techniques}
    \end{subfigure}
    
    \begin{subfigure}{\textwidth}
        \centering
        \includegraphics[width=.95\textwidth, clip, trim=15 0 40 0]{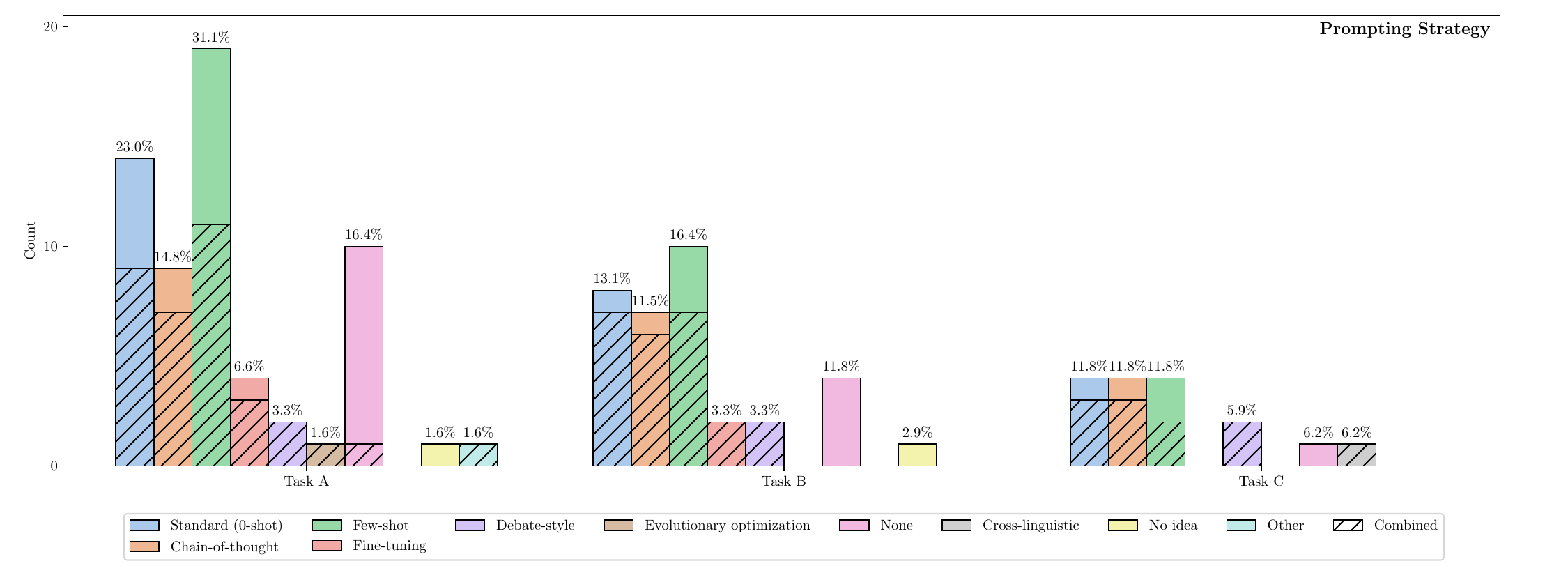}
        \label{fig:prompt_techniques}
    \end{subfigure}

    \caption{Overview of the most common architectures, training techniques, fine-tuning strategies, and prompting strategies used by our participants.}
    \label{fig:methods_overview}
\end{figure*}

\begin{figure*}
    \centering
    \begin{subfigure}{\textwidth}
        \centering
        \includegraphics[width=\textwidth]{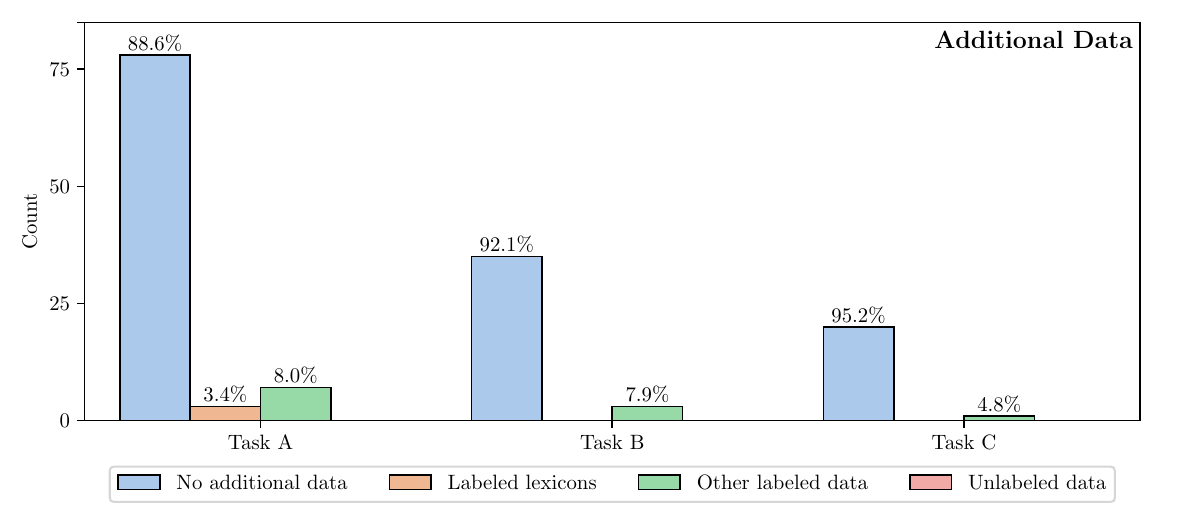}
        \label{fig:additional_data}
    \end{subfigure}
    
    \begin{subfigure}{\textwidth}
        \centering
        \includegraphics[width=\textwidth]{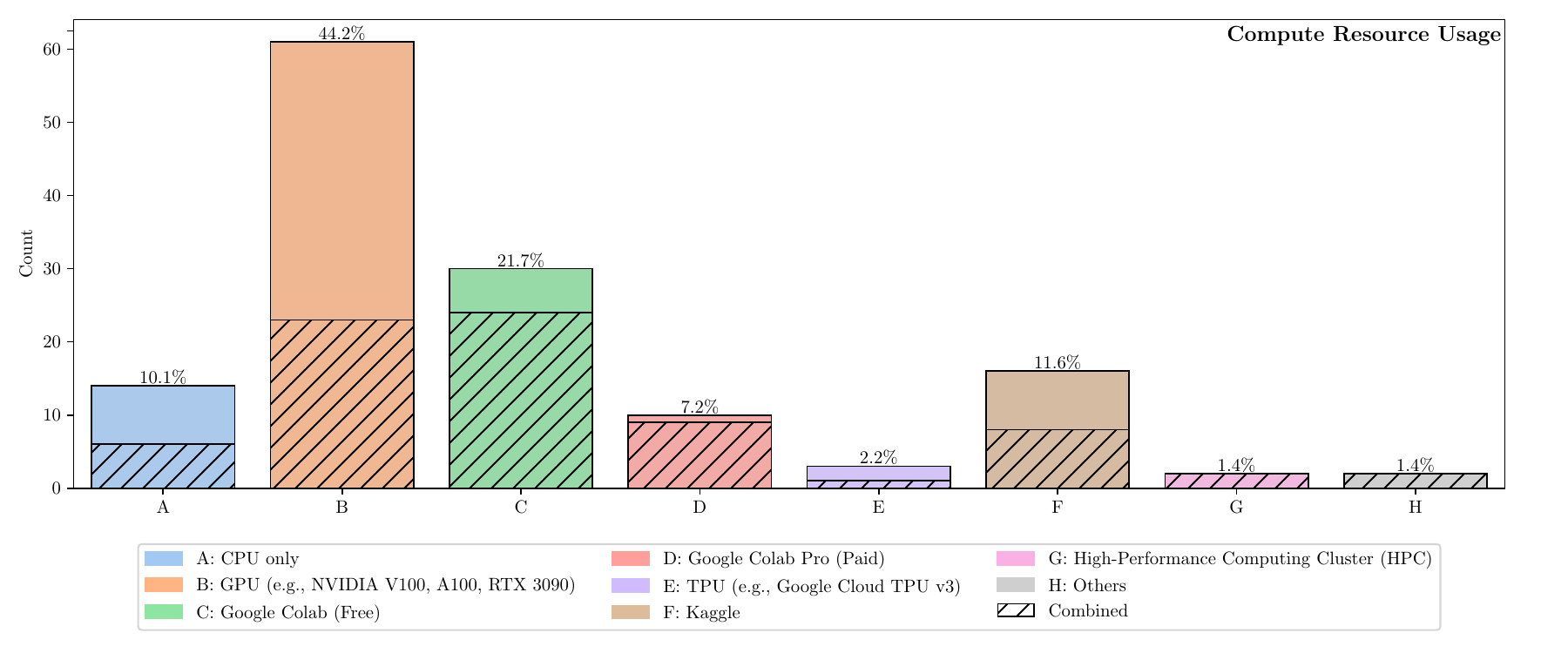}
    \label{fig:compute_resource}
    \end{subfigure}

    \caption{Overview of additional data and compute resources used by our participants.}
    \label{fig:data_compute_overview}
\end{figure*}

In this section, we provide the complete shared task results in \Cref{tab:track_a,,tab:track_b,,tab:track_c}, and some statistics on the architectures, training strategies, fine-tuning and prompting techniques used by our participants (\Cref{fig:methods_overview}). We also provide information on additional training data usage \Cref{fig:additional_data}.
\label{sec:appendix_results}

\onecolumn

\begin{landscape}
    \extrasmall
    \rowcolors{1}{white}{light-gray}
        \begin{longtable}{Rp{3cm}RRRRRRRRRRRRRRRRRRRRRRRRRRRRR}
        \hiderowcolors
            \caption{Track A Results} \label{tab:track_a} \\
            \toprule
            \textbf{S/N} & \textbf{Team Name} & \textbf{\texttt{afr}} & \textbf{\texttt{amh}} & \textbf{\texttt{arq}} & \textbf{\texttt{ary}} & \textbf{\texttt{chn}} & \textbf{\texttt{deu}} & \textbf{\texttt{eng}} & \textbf{\texttt{esp}} & \textbf{\texttt{hau}} & \textbf{\texttt{hin}} & \textbf{\texttt{ibo}} & \textbf{\texttt{kin}} & \textbf{\texttt{mar}} & \textbf{\texttt{orm}} & \textbf{\texttt{pcm}} & \textbf{\texttt{ptbr}} & \textbf{\texttt{ptmz}} & \textbf{\texttt{ron}} & \textbf{\texttt{rus}} & \textbf{\texttt{som}} & \textbf{\texttt{sun}} & \textbf{\texttt{swa}} & \textbf{\texttt{swe}} & \textbf{\texttt{tat}} & \textbf{\texttt{tir}} & \textbf{\texttt{ukr}} & \textbf{\texttt{vmw}} & \textbf{\texttt{yor}} & \textbf{avg} \\
            \midrule
            \endfirsthead

            \multicolumn{31}{l}{\normalsize\textbf{Table \thetable .} Continued from previous page} \\
            \toprule
            \extrasmall
            \textbf{S/N} & \textbf{Team Name} & \textbf{\texttt{afr}} & \textbf{\texttt{amh}} & \textbf{\texttt{arq}} & \textbf{\texttt{ary}} & \textbf{\texttt{chn}} & \textbf{\texttt{deu}} & \textbf{\texttt{eng}} & \textbf{\texttt{esp}} & \textbf{\texttt{hau}} & \textbf{\texttt{hin}} & \textbf{\texttt{ibo}} & \textbf{\texttt{kin}} & \textbf{\texttt{mar}} & \textbf{\texttt{orm}} & \textbf{\texttt{pcm}} & \textbf{\texttt{ptbr}} & \textbf{\texttt{ptmz}} & \textbf{\texttt{ron}} & \textbf{\texttt{rus}} & \textbf{\texttt{som}} & \textbf{\texttt{sun}} & \textbf{\texttt{swa}} & \textbf{\texttt{swe}} & \textbf{\texttt{tat}} & \textbf{\texttt{tir}} & \textbf{\texttt{ukr}} & \textbf{\texttt{vmw}} & \textbf{\texttt{yor}} & \textbf{avg} \\
            \midrule
            \endhead

            \midrule
            \multicolumn{31}{l}{\textbf{Continued on next page}} \\
            \midrule
            \endfoot

            \bottomrule
            \endlastfoot
            \showrowcolors

1 & PAI & \cellcolor{blue!30}69.9 & 64.7 & \cellcolor{blue!30}66.9 & \cellcolor{blue!30}62.9 & \cellcolor{blue!30}70.9 & \cellcolor{blue!30}74.0 & \cellcolor{blue!30}82.3 & \cellcolor{blue!30}84.9 & \cellcolor{blue!30}75.1 & 92.0 & \cellcolor{blue!30}60.0 & \cellcolor{blue!30}65.7 & \cellcolor{blue!30}88.4 & 58.2 & \cellcolor{blue!30}67.4 & \cellcolor{blue!30}68.3 & \cellcolor{blue!30}54.8 & \cellcolor{blue!30}79.4 & 88.2 & \cellcolor{blue!30}57.6 & 54.1 & 38.5 & \cellcolor{blue!30}62.6 & \cellcolor{blue!30}84.6 & 53.8 & \cellcolor{blue!30}72.6 & 25.5 & \cellcolor{blue!30}46.1 & 66.8 \\
2 & University of Indonesia & 54.6 & 51.2 & 55.0 & 53.5 & 62.9 & 66.2 & 74.9 & 79.5 & 64.4 & 86.1 & 49.6 & 48.4 & 84.7 & 46.2 & 60.5 & 56.9 & 42.7 & 73.8 & 84.4 & 43.7 & 43.2 & 37.9 & 57.8 & 60.4 & 40.0 & 63.4 & 13.6 & 29.0 & 56.6 \\
3 & LATE-GIL-NLP & 35.0 & 58.1 & 48.6 & 51.3 & 57.4 & 63.3 & 76.3 & 75.2 & 61.5 & 87.3 & 56.3 & 51.1 & 78.8 & 59.2 & 51.7 & 54.7 & 47.5 & 70.8 & 79.8 & 36.9 & 40.8 & 28.5 & 46.0 & 65.5 & 58.7 & 47.5 & 16.3 & 37.5 & 55.1 \\
4 & Zero\_Shot & 27.0 & 54.1 & 23.9 & 23.1 & 46.7 & 46.7 & 69.5 & 73.6 & 55.8 & 83.8 & 47.0 & 19.8 & 80.7 & 28.6 & 42.9 & 31.3 & 12.6 & 64.5 & 84.3 & 30.9 & 26.5 & 15.8 & 44.5 & 40.6 & 31.3 & 44.6 & 0.0 & 12.5 & 41.5 \\
5 & NTA & 40.7 & 67.2 & 50.6 & 24.9 & 59.4 & 57.1 & 76.1 & 75.3 & 67.8 & 83.7 & 47.0 & 38.0 & 81.4 & 50.5 & 47.8 & 34.1 & 34.6 & 70.0 & 83.5 & 47.1 & 42.0 & 21.4 & 52.9 & 56.9 & \cellcolor{blue!30}59.1 & 53.3 & 20.8 & 21.9 & 52.3 \\
6 & XLM-Muriel & 42.7 & 66.4 & 46.1 & 50.9 & 59.3 & 58.4 & 62.8 & 74.6 & 58.8 & 84.0 & 47.8 & 32.1 & 84.2 & 50.8 & 53.1 & 51.6 & 43.2 & 65.9 & 84.2 & 44.6 & 44.1 & 27.1 & 58.6 & 67.5 & 48.2 & 61.9 & 16.2 & 30.2 & 54.1 \\
7 & PromotionGo & 22.1 & 42.7 & 40.7 & 37.9 & 39.4 & 43.2 & 65.5 & 60.1 & 53.4 & 66.4 & 47.6 & 31.3 & 64.5 & 45.8 & 38.6 & 32.8 & 30.9 & 61.7 & 62.7 & 35.2 & 32.1 & 15.1 & 36.8 & 52.0 & 38.3 & 31.8 & 13.0 & 21.1 & 41.5 \\
8 & OZemi & 46.9 & 37.0 & 48.0 & 34.0 & 49.9 & 50.8 & 63.8 & 52.5 & 37.6 & 46.8 & 28.5 & 29.9 & 49.8 & 31.1 & 46.4 & 34.6 & 24.2 & 64.2 & 70.6 & 30.9 & 39.9 & 23.4 & 38.3 & 40.6 & 33.1 & 27.9 & 19.3 & 21.1 & 40.0 \\
9 & INFOTEC-NLP & 50.9 & 69.4 & 51.7 & 55.5 & 64.8 & 66.3 & 70.9 & 79.2 & 64.5 & 89.4 & 50.0 & 35.3 & 88.0 & 55.1 & 57.9 & 55.9 & 48.5 & 74.0 & 87.2 & 48.9 & 42.9 & 28.1 & 59.4 & 68.1 & 49.3 & 66.0 & 4.0 & 15.3 & 57.0 \\
10 & Pixel Phantoms & 30.6 & 43.7 & 32.1 & 34.2 & 39.6 & 37.9 & 59.7 & 64.0 & 50.1 & 73.0 & 41.0 & 31.7 & 70.2 & 31.9 & 40.9 & 27.5 & 20.8 & 57.6 & 72.0 & 25.4 & 30.9 & 17.8 & 38.9 & 40.6 & 32.0 & 35.3 & 7.8 & 14.7 & 39.4 \\
11 & YNWA\_PZ & 54.0 & 61.2 & 51.1 & 51.9 & 56.6 & 60.6 & 74.0 & 76.2 & 63.2 & 80.3 & 50.9 & 51.9 & 81.1 & 54.3 & 53.1 & 48.0 & 50.1 & 73.8 & 82.4 & 48.3 & 42.5 & 29.5 & 56.5 & 64.3 & 52.4 & 48.6 & 16.8 & 34.1 & 56.0 \\
12 & Deepwave & 52.3 & 50.7 & 48.2 & 52.4 & 62.0 & 65.5 & 75.1 & 79.4 & 59.4 & 90.1 & 48.1 & 38.0 & 87.8 & 44.6 & 59.5 & 53.9 & 45.1 & 70.3 & 87.2 & 30.9 & 37.9 & 22.7 & 56.9 & 68.2 & 40.9 & 61.9 & 7.8 & 24.0 & 54.3 \\
13 & UB\_Tel-U & 55.1 & 48.4 & 52.9 & 43.3 & 50.6 & 58.3 & 70.3 & 70.8 & 51.6 & 66.7 & 38.6 & 32.9 & 68.4 & 35.9 & 53.8 & 47.1 & 36.7 & 69.2 & 65.5 & 29.9 & 41.9 & 30.2 & 50.6 & 54.6 & 40.5 & 42.1 & 14.0 & 22.5 & 47.9 \\
14 & Heimerdinger & 57.4 & 55.8 & 59.5 & 58.2 & 68.0 & 70.6 & 80.4 & 83.8 & 61.5 & 89.6 & 54.1 & 44.5 & 87.7 & 50.9 & 60.5 & 62.5 & 50.7 & 74.6 & \cellcolor{blue!30}90.1 & 43.8 & 48.2 & 30.2 & 58.1 & 75.4 & 45.9 & 63.7 & 22.2 & 39.2 & 60.3 \\
15 & Chinchunmei & 57.7 & \cellcolor{blue!30}77.3 & 57.2 & 56.3 & 66.8 & 70.4 & 79.7 & 81.7 & 67.6 & 90.0 & 55.9 & 47.0 & 87.2 & 53.7 & 63.3 & 60.3 & 50.3 & 74.7 & 88.7 & 43.4 & 45.4 & 31.9 & 58.5 & 75.8 & 48.0 & 65.8 & 19.7 & 28.9 & 60.8 \\
16 & TeleAI & 57.8 & 37.6 & 39.8 & 52.8 & 68.2 & 67.2 & 80.6 & 80.7 & 53.8 & 85.2 & 38.5 & 39.5 & 78.3 & 38.1 & 54.6 & 59.9 & 44.9 & 68.2 & 87.7 & 35.2 & 46.6 & 32.0 & 57.3 & 52.9 & 37.0 & 62.3 & 9.0 & 32.2 & 53.5 \\
17 & FiRC-NLP & 58.0 & 69.6 & 55.2 & 58.5 & 61.7 & 68.4 & 74.2 & 79.7 & 69.4 & 87.8 & 54.1 & 41.5 & 86.7 & 60.7 & 61.1 & 60.1 & 51.1 & 74.9 & 86.6 & 55.9 & 49.8 & 32.3 & 60.9 & 71.7 & 57.6 & 67.3 & 25.0 & 30.8 & 61.1 \\
18 & maomao & 68.7 & 49.1 & 45.4 & 54.8 & 60.3 & 57.0 & 63.2 & 77.8 & 55.7 & 81.9 & 32.0 & 43.1 & 83.7 & 28.4 & 54.7 & 61.2 & 46.4 & 71.1 & 83.1 & 42.3 & 45.3 & 31.3 & 56.0 & 72.8 & 38.0 & 63.3 & 16.7 & 34.3 & 54.2 \\
19 & McGill-NLP & 60.1 & 63.4 & 58.7 & 52.5 & 59.6 & 64.4 & 77.7 & 79.0 & 65.4 & 88.0 & 53.2 & 59.0 & 87.5 & 56.4 & 63.1 & 56.8 & 45.1 & 72.8 & 86.4 & 48.4 & 50.3 & 35.9 & 51.1 & 73.9 & 48.2 & 60.0 & 12.6 & 33.0 & 59.4 \\
20 & Empaths & 66.7 & 55.8 & 52.0 & 37.0 & 62.5 & 21.6 & 78.8 & 81.7 & 69.5 & 86.6 & 53.8 & 56.3 & 86.4 & 49.0 & 59.9 & 37.2 & 49.3 & 76.1 & 87.9 & 50.8 & 52.9 & \cellcolor{blue!30}38.6 & 49.3 & 77.6 & 20.4 & 61.6 & 21.6 & 38.2 & 56.4 \\
21 & Pateam & 26.6 & 33.3 & 56.3 & 44.5 & 58.0 & 43.8 & 80.5 & 82.8 & 36.7 & 78.2 & 26.0 & 21.9 & 67.5 & 40.6 & 33.7 & 42.7 & 49.2 & 41.3 & 61.3 & 21.1 & 42.3 & - & 37.6 & 39.7 & 10.5 & 50.4 & 6.7 & 13.9 & 42.5 \\
22 & UoB-NLP & 45.0 & 67.0 & 46.9 & 51.5 & 56.9 & 54.8 & 64.5 & 72.9 & 66.3 & 84.2 & 47.2 & 51.6 & 80.0 & 52.6 & 51.5 & 47.8 & 40.0 & 68.1 & - & 44.5 & 36.5 & 26.2 & 52.1 & 63.7 & 50.3 & 58.4 & 5.6 & 19.3 & 52.1 \\
23 & YNU-NPCC & 43.5 & 47.6 & 44.4 & 48.4 & 55.8 & 60.3 & - & 73.1 & 50.0 & 82.2 & 36.6 & 44.3 & 82.0 & 40.1 & 44.5 & 53.2 & 38.6 & 60.2 & 79.9 & 38.2 & 49.0 & 31.3 & 52.2 & 70.5 & 28.2 & 56.9 & 4.6 & 26.1 & 49.7 \\
24 & JNLP & 59.2 & 67.7 & 64.1 & 60.9 & 68.0 & - & 80.4 & 83.0 & 65.0 & \cellcolor{blue!30}92.6 & 54.0 & 42.9 & 87.8 & 57.3 & 63.4 & 61.8 & 45.4 & 77.9 & 89.1 & 49.6 & 46.0 & 29.5 & 61.9 & 72.2 & 48.5 & - & 22.6 & 36.1 & 61.0 \\
25 & CSIRO-LT & 31.6 & 58.8 & 52.5 & - & 57.8 & 65.7 & 77.6 & 82.0 & 54.2 & 73.2 & 32.6 & 43.0 & 75.8 & 39.2 & 21.8 & 55.5 & - & 72.2 & 89.1 & 43.3 & 48.8 & 30.3 & 48.9 & 53.8 & 50.0 & 66.4 & 17.3 & 30.0 & 52.7 \\
26 & Howard University-AI4PC & - & 15.4 & 30.4 & 32.3 & 39.3 & 40.1 & - & 57.0 & 31.8 & 62.9 & 18.5 & 16.6 & 58.4 & 17.5 & 29.5 & 38.6 & 28.2 & 43.8 & 59.0 & 16.0 & 32.3 & 22.9 & 30.7 & 30.1 & - & 35.3 & 12.8 & 13.6 & 32.5 \\
27 & Tue-JMS & - & 69.6 & 53.4 & 58.0 & 60.3 & 64.5 & 68.5 & 79.4 & 69.0 & 88.5 & 53.0 & 53.2 & 87.3 & 50.9 & 56.9 & 56.5 & 47.8 & 73.7 & 88.0 & 47.8 & 43.9 & - & 59.0 & 79.7 & 53.3 & 63.4 & - & - & 63.6 \\
28 & SyntaxMind & 36.5 & - & 45.7 & 37.3 & 55.8 & 48.7 & 66.5 & 57.4 & - & 65.1 & - & - & 72.5 & - & - & 31.4 & 37.1 & 61.7 & 66.0 & - & 35.6 & 24.1 & 43.3 & 49.1 & - & 31.5 & - & 26.1 & 46.9 \\
29 & IASBS & 59.1 & 52.3 & - & - & 55.4 & 61.4 & 73.5 & 72.8 & 65.8 & 81.1 & 54.4 & - & 80.1 & - & - & - & - & 72.5 & 82.2 & - & - & - & 53.0 & - & - & 58.8 & - & - & 65.9 \\
30 & Emotion Train & - & - & 56.5 & 56.8 & - & 65.2 & 75.6 & 76.1 & - & - & - & - & - & - & 59.3 & 53.6 & 47.1 & 72.2 & 88.2 & - & - & 12.4 & 53.2 & 49.8 & - & 45.3 & - & - & 58.0 \\
31 & Team A & - & 68.5 & - & 55.5 & 61.2 & 66.5 & 73.4 & 80.5 & 67.7 & 89.0 & - & - & 86.6 & 56.3 & - & - & - & 74.0 & 88.3 & - & - & - & - & - & - & 60.1 & - & - & 71.4 \\
32 & Habib University & 25.7 & 45.5 & - & 14.5 & - & 44.6 & 71.0 & 71.3 & - & 80.8 & - & - & 76.4 & - & - & - & - & 50.4 & 72.1 & - & - & - & 21.9 & - & - & 29.1 & - & - & 50.3 \\
33 & JellyK & - & - & - & - & 63.9 & 65.4 & 76.2 & 77.2 & - & 86.1 & - & - & 86.3 & - & 58.2 & - & 42.2 & 74.2 & 88.9 & - & - & - & - & 59.3 & - & 57.9 & - & - & 69.6 \\
34 & CSECU-Learners & - & 70.2 & 55.5 & - & 60.0 & 60.2 & 73.8 & 76.9 & 67.3 & - & - & - & - & - & - & 52.4 & - & 74.7 & 84.7 & - & - & - & - & - & - & 50.6 & - & - & 66.0 \\
35 & UncleLM & - & - & - & 53.9 & 61.5 & 58.2 & 74.6 & 76.4 & 31.8 & 85.6 & - & - & 80.0 & - & - & - & 44.0 & 74.5 & 87.1 & - & - & - & - & - & - & - & - & - & 66.1 \\
36 & Domain\_adaptation & - & 47.8 & 26.4 & - & 57.2 & 57.7 & - & 72.0 & 30.2 & - & - & - & - & - & - & 49.3 & - & 54.0 & 76.0 & - & - & - & - & - & - & 50.4 & - & - & 52.1 \\
37 & Team KiAmSo & 46.7 & 53.4 & 49.1 & 50.1 & 56.6 & 59.8 & 72.5 & 76.3 & 58.4 & 78.5 & - & - & - & - & - & - & - & - & - & - & - & - & - & - & - & - & - & - & 60.1 \\
38 & EMO-NLP & - & 60.2 & - & - & - & 53.1 & 66.8 & - & - & - & - & - & - & 31.8 & - & 45.9 & - & - & 75.9 & 37.8 & 29.3 & - & - & - & 32.5 & - & - & - & 48.1 \\
39 & GT-NLP & 51.9 & - & 52.7 & 53.1 & - & 66.3 & - & - & - & - & - & - & - & - & - & 59.9 & - & - & 86.4 & - & - & - & 56.4 & - & - & 60.0 & - & - & 60.8 \\
40 & Trans-Sent & - & 54.7 & - & - & - & 48.0 & 70.4 & - & - & 82.4 & - & - & 80.3 & - & - & - & - & 70.9 & 88.3 & - & - & - & - & - & - & - & - & - & 70.7 \\
41 & NUST Titans & - & 71.4 & - & - & - & - & - & - & 69.2 & - & 18.1 & - & - & - & - & - & - & - & - & - & 34.5 & 29.5 & - & - & - & - & - & 31.2 & 42.3 \\
42 & Zero & - & - & - & - & - & - & 71.7 & 35.9 & - & - & - & - & - & - & - & - & - & 36.7 & 52.9 & - & - & - & - & - & - & - & - & - & 49.3 \\
43 & Tewodros & - & 70.0 & - & - & - & - & 71.2 & 79.4 & - & - & - & - & - & \cellcolor{blue!30}61.6 & - & - & - & - & - & - & - & - & - & - & - & - & - & - & 70.6 \\
44 & UIMP-Aaman & - & - & - & - & - & 62.4 & 72.2 & 76.7 & - & - & - & - & - & - & - & 51.7 & - & - & - & - & - & - & - & - & - & - & - & - & 65.8 \\
45 & GinGer & 45.4 & - & 54.7 & - & - & - & - & - & - & 83.9 & - & - & - & - & - & - & - & - & - & - & - & - & 52.3 & - & - & - & - & - & 59.1 \\
46 & DUT\_IR & - & - & - & - & - & - & 81.2 & - & - & 84.2 & - & - & 83.7 & - & - & - & - & - & 74.4 & - & - & - & - & - & - & - & - & - & 80.9 \\
47 & DataBees & - & - & - & - & - & 51.7 & 37.2 & 69.6 & - & - & - & - & - & - & - & - & - & - & - & - & - & - & - & - & - & - & - & - & 52.8 \\
48 & Team Unibuc & - & - & - & - & - & - & 75.5 & - & - & - & - & - & - & - & - & - & 17.3 & - & - & - & - & - & - & - & - & - & \cellcolor{blue!30}32.5 & - & 41.8 \\
49 & indiDataMiner & - & - & - & - & - & - & 66.3 & - & - & 87.4 & - & - & 88.3 & - & - & - & - & - & - & - & - & - & - & - & - & - & - & - & 80.7 \\
50 & CIOL & - & - & - & - & - & - & 62.1 & - & - & - & - & - & - & - & - & 27.7 & - & - & 84.8 & - & - & - & - & - & - & - & - & - & 58.2 \\
51 & CIC-IPN & - & - & - & - & - & - & 70.6 & 73.2 & - & - & - & - & - & - & - & - & - & - & - & - & - & - & - & - & - & - & - & 28.2 & 57.3 \\
52 & Team UBD & - & - & - & - & - & 54.3 & 68.0 & 76.2 & - & - & - & - & - & - & - & - & - & - & - & - & - & - & - & - & - & - & - & - & 66.2 \\
53 & ITF-NLP & - & - & - & - & - & 62.2 & 75.0 & - & - & - & - & - & - & - & - & - & - & - & - & - & - & - & - & - & - & - & - & - & 68.6 \\
54 & IEGPS-CSIC & - & - & - & - & - & - & 71.0 & - & - & - & - & - & - & - & - & - & - & - & 84.2 & - & - & - & - & - & - & - & - & - & 77.6 \\
55 & NLP-DU & - & - & - & - & - & - & 70.6 & - & - & - & - & - & - & - & - & - & - & - & - & - & - & - & - & - & - & - & - & - & 70.6 \\
56 & zhouyijiang1 & - & - & - & - & - & - & 70.5 & - & - & - & - & - & - & - & - & - & - & - & - & - & - & - & - & - & - & - & - & - & 70.5 \\
57 & wangkongqiang & - & - & - & - & - & - & 70.0 & - & - & - & - & - & - & - & - & - & - & - & - & - & - & - & - & - & - & - & - & - & 70.0 \\
58 & JUNLP & - & - & - & - & - & - & 69.9 & - & - & - & - & - & - & - & - & - & - & - & - & - & - & - & - & - & - & - & - & - & 69.9 \\
59 & dianchi & - & - & - & - & - & - & 69.6 & - & - & - & - & - & - & - & - & - & - & - & - & - & - & - & - & - & - & - & - & - & 69.6 \\
60 & RSSN & - & - & - & - & - & - & 69.4 & - & - & - & - & - & - & - & - & - & - & - & - & - & - & - & - & - & - & - & - & - & 69.4 \\
61 & CodeBlAIzers & - & - & - & - & - & - & 55.2 & - & - & - & - & - & - & - & - & - & - & - & - & - & - & - & - & - & - & - & - & - & 55.2 \\
62 & Angeliki Linardatou & - & - & - & - & - & - & 67.8 & - & - & - & - & - & - & - & - & - & - & - & - & - & - & - & - & - & - & - & - & - & 67.8 \\
63 & TILeN & - & - & - & - & - & - & - & - & - & - & - & - & - & - & - & - & - & - & 82.1 & - & - & - & - & - & - & - & - & - & 82.1 \\
64 & AKCIT & - & - & - & - & - & - & - & - & - & - & - & - & - & - & - & 60.2 & - & - & - & - & - & - & - & - & - & - & - & - & 60.2 \\
65 & STFXNLP & - & - & - & - & - & - & 64.7 & - & - & - & - & - & - & - & - & - & - & - & - & - & - & - & - & - & - & - & - & - & 64.7 \\
66 & VerbaNexAI & - & - & - & - & - & - & 62.7 & - & - & - & - & - & - & - & - & - & - & - & - & - & - & - & - & - & - & - & - & - & 62.7 \\
67 & NLP-Cimat & - & - & - & - & - & - & 62.4 & - & - & - & - & - & - & - & - & - & - & - & - & - & - & - & - & - & - & - & - & - & 62.4 \\
68 & HausaNLP & - & - & - & - & - & - & - & - & 56.0 & - & - & - & - & - & - & - & - & - & - & - & - & - & - & - & - & - & - & - & 56.0 \\
69 & TechSSN3 & - & - & - & - & - & - & 69.0 & - & - & - & - & - & - & - & - & - & - & - & - & - & - & - & - & - & - & - & - & - & 69.0 \\
70 & HTU & - & - & 48.6 & - & - & - & - & - & - & - & - & - & - & - & - & - & - & - & - & - & - & - & - & - & - & - & - & - & 48.6 \\
71 & xiacui & - & - & - & - & - & - & 72.2 & - & - & - & - & - & - & - & - & - & - & - & - & - & - & - & - & - & - & - & - & - & 72.2 \\
72 & GOLDX & - & - & - & - & - & - & 76.2 & - & - & - & - & - & - & - & - & - & - & - & - & - & - & - & - & - & - & - & - & - & 76.2 \\
73 & Tarbiatmodares & - & 65.5 & - & - & - & - & - & - & - & - & - & - & - & - & - & - & - & - & - & - & - & - & - & - & - & - & - & - & 65.5 \\
74 & Amado & - & 63.0 & - & - & - & - & - & - & - & - & - & - & - & - & - & - & - & - & - & - & - & - & - & - & - & - & - & - & 63.0 \\
75 & AtlasIA & - & - & - & 33.5 & - & - & - & - & - & - & - & - & - & - & - & - & - & - & - & - & - & - & - & - & - & - & - & - & 33.5 \\
76 & NYCU-NLP & - & - & - & - & - & - & 82.2 & - & - & - & - & - & - & - & - & - & - & - & - & - & - & - & - & - & - & - & - & - & 82.2 \\
77 & NCL-NLP & - & - & - & - & - & - & 78.6 & - & - & - & - & - & - & - & - & - & - & - & - & - & - & - & - & - & - & - & - & - & 78.6 \\
78 & MRS & - & - & - & - & - & - & 76.6 & - & - & - & - & - & - & - & - & - & - & - & - & - & - & - & - & - & - & - & - & - & 76.6 \\
79 & NITK-VITAL & - & - & - & - & - & - & 75.6 & - & - & - & - & - & - & - & - & - & - & - & - & - & - & - & - & - & - & - & - & - & 75.6 \\
80 & CharsiuRice & - & - & - & - & - & - & 72.6 & - & - & - & - & - & - & - & - & - & - & - & - & - & - & - & - & - & - & - & - & - & 72.6 \\
81 & QiMP & - & - & - & - & - & - & 75.4 & - & - & - & - & - & - & - & - & - & - & - & - & - & - & - & - & - & - & - & - & - & 75.4 \\
82 & NLP\_goats & - & - & - & - & - & - & 75.0 & - & - & - & - & - & - & - & - & - & - & - & - & - & - & - & - & - & - & - & - & - & 75.0 \\
83 & AGHNA & - & - & - & - & - & - & 73.7 & - & - & - & - & - & - & - & - & - & - & - & - & - & - & - & - & - & - & - & - & - & 73.7 \\
84 & OPI-DRO-HEL at & - & - & - & - & - & - & 73.6 & - & - & - & - & - & - & - & - & - & - & - & - & - & - & - & - & - & - & - & - & - & 73.6 \\
85 & Exploration Lab IITK & - & - & - & - & - & - & 73.4 & - & - & - & - & - & - & - & - & - & - & - & - & - & - & - & - & - & - & - & - & - & 73.4 \\
86 & Lotus & - & - & - & - & - & - & 73.2 & - & - & - & - & - & - & - & - & - & - & - & - & - & - & - & - & - & - & - & - & - & 73.2 \\
87 & Lazarus NLP & - & - & - & - & - & - & - & - & - & - & - & - & - & - & - & - & - & - & - & - & \cellcolor{blue!30}55.0 & - & - & - & - & - & - & - & 55.0 \\
        \end{longtable}
\end{landscape}

\begin{table*}[]
    \centering
    \resizebox{\textwidth}{!}{
    \rowcolors{1}{white}{light-gray}
    \begin{tabular}{clrrrrrrrrrrrr}
\toprule
\textbf{S/N} & \textbf{Team Name} & \textbf{\texttt{amh}} & \textbf{\texttt{arq}} & \textbf{\texttt{chn}} & \textbf{\texttt{deu}} & \textbf{\texttt{eng}} & \textbf{\texttt{esp}} & \textbf{\texttt{hau}} & \textbf{\texttt{ptbr}} & \textbf{\texttt{ron}} & \textbf{\texttt{rus}} & \textbf{\texttt{ukr}} & \textbf{avg} \\
\midrule
1 & CSECU-Learners & \cellcolor{blue!30}85.6 & 44.3 & 57.1 & 53.3 & 65.0 & 71.5 & 65.6 & 46.6 & 63.7 & 83.3 & 47.8 & 62.2 \\
2 & Heimerdinger & 78.0 & 50.7 & 67.1 & 72.3 & 79.3 & 78.8 & 61.8 & 65.1 & 65.7 & 88.3 & 60.3 & 69.8 \\
3 & CSIRO-LT & 15.4 & 52.1 & 54.9 & 61.0 & 72.1 & 69.7 & 37.1 & 48.9 & 59.2 & 79.3 & 49.4 & 54.5 \\
4 & TeleAI & 32.2 & 52.4 & 70.8 & 74.2 & 83.2 & 78.6 & 57.9 & 69.0 & 70.4 & 91.8 & 64.5 & 67.7 \\
5 & Pixel Phantoms & 37.7 & 11.2 & 36.6 & 34.2 & 32.9 & 52.8 & 46.7 & 25.0 & 40.6 & 74.5 & 24.8 & 37.9 \\
6 & IASBS & 46.1 & 49.8 & 53.6 & 59.8 & 69.3 & 69.2 & 65.8 & 52.1 & 62.0 & 86.0 & 56.5 & 60.9 \\
7 & maomao & 46.2 & 56.7 & 58.6 & 65.4 & 75.2 & 74.9 & 52.0 & 60.2 & 65.3 & 84.7 & 53.4 & 63.0 \\
8 & YNWA\_PZ & 49.4 & 36.5 & 48.5 & 54.1 & 68.8 & 66.7 & 58.4 & 38.2 & 57.6 & 78.4 & 42.5 & 54.5 \\
9 & UB\_Tel-U & 56.9 & 33.2 & 56.6 & 55.1 & 53.1 & 68.3 & 53.0 & 47.8 & 55.6 & 78.2 & 53.6 & 55.6 \\
10 & Habib University & 60.3 & 30.8 & 47.8 & 53.2 & 74.0 & 64.2 & 39.5 & 38.6 & 49.9 & 76.4 & 42.5 & 52.5 \\
11 & JNLP & 60.4 & 58.7 & 65.9 & 72.5 & 81.3 & 77.5 & 65.0 & 65.1 & 70.5 & 90.7 & 67.2 & 70.4 \\
12 & PAI & 64.6 & \cellcolor{blue!30}65.0 & \cellcolor{blue!30}72.2 & \cellcolor{blue!30}76.6 & \cellcolor{blue!30}84.0 & \cellcolor{blue!30}80.8 & \cellcolor{blue!30}77.0 & \cellcolor{blue!30}71.0 & \cellcolor{blue!30}72.6 & \cellcolor{blue!30}92.5 & \cellcolor{blue!30}70.8 & 75.2 \\
13 & Tue-JMS & 67.2 & 42.5 & 61.3 & 64.3 & 66.5 & 73.9 & 67.0 & 56.0 & 65.4 & 88.6 & 56.1 & 64.4 \\
14 & Deepwave & 68.0 & 57.4 & 69.4 & 74.2 & 80.9 & 79.2 & 74.7 & 68.4 & 71.6 & 91.6 & 66.2 & 72.9 \\
15 & Chinchunmei & 71.6 & 48.6 & 66.4 & 70.8 & 81.0 & 77.0 & 67.9 & 62.2 & 66.4 & 89.8 & 63.1 & 69.5 \\
16 & Domain\_adaptation & 41.3 & 20.6 & 3.4 & 30.1 & - & 45.5 & 54.6 & 23.9 & 36.8 & 56.1 & 21.1 & 33.3 \\
17 & GT-NLP & - & 52.8 & 57.7 & 49.1 & 75.1 & 75.8 & 63.3 & 61.6 & 68.5 & 88.0 & 31.9 & 62.4 \\
18 & SyntaxMind & - & 15.8 & 47.9 & 38.9 & 55.4 & 39.2 & - & 23.6 & 36.8 & 52.6 & 19.1 & 36.6 \\
19 & Zero\_Shot & - & 29.2 & 46.7 & 42.3 & - & 63.6 & 57.3 & - & 51.8 & 83.7 & - & 53.5 \\
20 & Stanford MLab & - & 21.9 & - & - & 69.0 & - & 54.0 & - & 39.4 & 75.9 & - & 52.0 \\
21 & UT-NLP & - & - & 30.7 & - & 56.9 & - & 33.7 & - & - & - & 35.2 & 39.1 \\
22 & UIMP-Aaman & - & - & - & 61.6 & 73.3 & 70.5 & - & 50.4 & - & - & - & 64.0 \\
23 & Tewodros & 65.0 & - & - & - & 59.4 & 62.3 & - & - & - & - & - & 62.2 \\
24 & CIOL & - & - & 48.3 & - & - & 68.5 & - & - & - & 85.9 & - & 67.6 \\
25 & Team UBD & - & - & - & 48.9 & - & 70.6 & - & - & - & - & - & 59.8 \\
26 & Zero & - & - & - & - & 61.3 & - & - & - & - & - & - & 61.3 \\
27 & AKCIT & - & - & - & - & - & - & - & 64.8 & - & - & - & 64.8 \\
28 & DataBees & - & - & - & - & 3.0 & - & - & - & - & - & - & 3.0 \\
29 & QiMP & - & - & - & - & 49.2 & - & - & - & - & - & - & 49.2 \\
30 & CharsiuRice & - & - & - & - & 56.6 & - & - & - & - & - & - & 56.6 \\
31 & OZemi & - & - & - & - & 61.2 & - & - & - & - & - & - & 61.2 \\
32 & tinaal & - & - & - & - & 67.0 & - & - & - & - & - & - & 67.0 \\
33 & CIC-IPN & - & - & - & - & 63.9 & - & - & - & - & - & - & 63.9 \\
34 & indiDataMiner & - & - & - & - & 69.0 & - & - & - & - & - & - & 69.0 \\
35 & NLP\_goats & - & - & - & - & 75.1 & - & - & - & - & - & - & 75.1 \\
36 & AGHNA & - & - & - & - & 76.2 & - & - & - & - & - & - & 76.2 \\
37 & NYCU-NLP & - & - & - & - & 83.7 & - & - & - & - & - & - & 83.7 \\
38 & TILeN & - & - & - & - & - & - & - & - & - & 80.7 & - & 80.7 \\
\bottomrule
\end{tabular}
}
    \caption{Track B Results}
    \label{tab:track_b}
\end{table*}

\begin{landscape}
    \extraextrasmall
    \rowcolors{1}{white}{light-gray}
        \begin{longtable}{Lp{2.6cm}LLLLLLLLLLLLLLLLLLLLLLLLLLLLLLLLL}
        \hiderowcolors
            \caption{Track C Results} \label{tab:track_c} \\
            \toprule
            \textbf{S/N} & \textbf{Team Name} & \textbf{\texttt{afr}} & \textbf{\texttt{amh}} & \textbf{\texttt{arq}} & \textbf{\texttt{ary}} & \textbf{\texttt{chn}} & \textbf{\texttt{deu}} & \textbf{\texttt{eng}} & \textbf{\texttt{esp}} & \textbf{\texttt{hau}} & \textbf{\texttt{hin}} & \textbf{\texttt{ibo}} & \textbf{\texttt{ind}} & \textbf{\texttt{jav}} & \textbf{\texttt{kin}} & \textbf{\texttt{mar}} & \textbf{\texttt{orm}} & \textbf{\texttt{pcm}} & \textbf{\texttt{ptbr}} & \textbf{\texttt{ptmz}} & \textbf{\texttt{ron}} & \textbf{\texttt{rus}} & \textbf{\texttt{som}} & \textbf{\texttt{sun}} & \textbf{\texttt{swa}} & \textbf{\texttt{swe}} & \textbf{\texttt{tat}} & \textbf{\texttt{tir}} & \textbf{\texttt{ukr}} & \textbf{\texttt{vmw}} & \textbf{\texttt{xho}} & \textbf{\texttt{yor}} & \textbf{\texttt{zul}} & \textbf{avg} \\
            \midrule
            \endfirsthead

            \multicolumn{35}{l}{\textbf{Table \thetable .} Continued from previous page} \\
            \toprule
            \textbf{S/N} & \textbf{Team Name} & \textbf{\texttt{afr}} & \textbf{\texttt{amh}} & \textbf{\texttt{arq}} & \textbf{\texttt{ary}} & \textbf{\texttt{chn}} & \textbf{\texttt{deu}} & \textbf{\texttt{eng}} & \textbf{\texttt{esp}} & \textbf{\texttt{hau}} & \textbf{\texttt{hin}} & \textbf{\texttt{ibo}} & \textbf{\texttt{ind}} & \textbf{\texttt{jav}} & \textbf{\texttt{kin}} & \textbf{\texttt{mar}} & \textbf{\texttt{orm}} & \textbf{\texttt{pcm}} & \textbf{\texttt{ptbr}} & \textbf{\texttt{ptmz}} & \textbf{\texttt{ron}} & \textbf{\texttt{rus}} & \textbf{\texttt{som}} & \textbf{\texttt{sun}} & \textbf{\texttt{swa}} & \textbf{\texttt{swe}} & \textbf{\texttt{tat}} & \textbf{\texttt{tir}} & \textbf{\texttt{ukr}} & \textbf{\texttt{vmw}} & \textbf{\texttt{xho}} & \textbf{\texttt{yor}} & \textbf{\texttt{zul}} & \textbf{avg} \\
            \midrule
            \endhead

            \midrule
            \multicolumn{35}{l}{\textbf{Continued on next page}} \\
            \midrule
            \endfoot

            \bottomrule
            \endlastfoot
            \showrowcolors

            1 & maomao & \cellcolor{blue!30}70.5 & 58.0 & 58.4 & 56.5 & 62.2 & 68.3 & 75.4 & 80.6 & 59.1 & 89.6 & 43.3 & \cellcolor{blue!30}67.2 & 42.2 & 44.9 & 86.3 & 36.0 & 56.2 & 61.7 & 49.5 & 74.7 & 85.2 & \cellcolor{blue!30}48.8 & 46.4 & \cellcolor{blue!30}38.0 & 57.8 & 69.7 & 37.9 & 62.3 & 17.3 & \cellcolor{blue!30}44.3 & \cellcolor{blue!30}35.9 & \cellcolor{blue!30}39.7 & 57.0 \\
2 & YNWA\_PZ & 54.0 & 61.2 & 51.1 & 51.9 & 56.6 & 60.6 & 74.0 & 76.2 & 63.2 & 80.3 & 50.9 & 35.6 & 25.6 & \cellcolor{blue!30}51.9 & 81.1 & \cellcolor{blue!30}54.3 & 53.1 & 48.0 & 50.1 & 73.8 & 82.4 & 48.3 & 42.5 & 29.5 & 56.5 & 64.3 & \cellcolor{blue!30}52.4 & 48.6 & 16.8 & 16.6 & 34.1 & 16.4 & 51.9 \\
3 & Deepwave & 57.4 & \cellcolor{blue!30}66.1 & \cellcolor{blue!30}58.8 & \cellcolor{blue!30}63.2 & \cellcolor{blue!30}68.9 & \cellcolor{blue!30}72.7 & \cellcolor{blue!30}79.7 & \cellcolor{blue!30}83.1 & \cellcolor{blue!30}70.9 & \cellcolor{blue!30}91.9 & \cellcolor{blue!30}60.5 & 55.4 & 37.2 & 50.8 & \cellcolor{blue!30}90.3 & 54.2 & \cellcolor{blue!30}67.4 & \cellcolor{blue!30}62.9 & \cellcolor{blue!30}55.5 & \cellcolor{blue!30}76.7 & \cellcolor{blue!30}90.6 & \cellcolor{blue!30}48.8 & \cellcolor{blue!30}46.7 & 35.5 & \cellcolor{blue!30}64.5 & \cellcolor{blue!30}78.9 & 50.5 & \cellcolor{blue!30}70.2 & \cellcolor{blue!30}21.0 & 15.9 & 34.2 & 19.3 & 59.4 \\
4 & OZemi & 47.1 & 37.0 & 47.9 & 33.9 & 49.9 & 50.8 & 63.8 & 52.7 & 37.8 & 46.7 & 27.6 & 48.8 & 41.3 & 29.9 & 49.7 & 31.2 & 46.4 & 34.5 & 24.2 & 64.5 & 44.0 & 31.0 & 40.8 & 23.2 & 38.8 & 40.4 & 33.1 & 28.1 & 19.3 & 31.5 & 21.1 & 21.2 & 38.7 \\
5 & UB\_Tel-U & 37.1 & 62.3 & 42.3 & 44.5 & 58.8 & 60.3 & 65.6 & 74.8 & 58.6 & 85.8 & 43.0 & 51.2 & 35.3 & 29.1 & 83.3 & 37.6 & 52.8 & 49.9 & 37.8 & 70.3 & 83.1 & 35.1 & 37.5 & 20.2 & 54.5 & 63.9 & 36.4 & 57.9 & 4.2 & 16.3 & 13.9 & 10.8 & 47.3 \\
6 & UoB-NLP & 36.3 & 62.7 & 44.1 & 42.4 & 56.9 & 55.4 & 64.5 & 72.9 & 62.7 & 80.0 & 48.4 & 33.3 & - & 46.6 & 80.0 & 49.1 & 51.3 & 44.0 & 32.5 & 68.1 & 80.6 & 39.7 & 29.6 & 22.0 & 48.3 & 63.7 & 44.5 & 49.8 & 3.4 & - & 19.3 & - & 49.4 \\
7 & CIOL & 15.7 & - & - & - & 29.9 & - & 22.3 & 26.2 & - & 22.7 & - & 26.1 & 24.4 & - & 17.9 & - & - & - & 9.8 & 23.1 & 31.0 & - & 22.1 & 19.0 & 21.3 & 23.0 & - & 19.7 & - & - & 15.1 & - & 21.7 \\
8 & IASBS & 32.8 & - & 48.5 & - & 54.3 & 47.2 & 73.0 & 69.6 & 59.8 & 76.4 & - & - & - & - & 77.4 & - & - & 41.0 & 43.5 & 64.9 & 81.7 & - & - & - & 47.3 & - & - & 50.9 & - & - & - & - & 57.9 \\
9 & Habib University & 30.0 & 27.0 & - & - & 39.1 & 8.0 & - & 24.1 & - & 67.4 & 9.2 & - & - & - & 76.4 & - & - & - & - & 47.6 & 20.7 & 27.9 & - & 11.6 & 22.1 & - & - & 17.9 & - & - & - & - & 30.6 \\
10 & Howard University-AI4PC & - & 14.0 & 41.8 & 29.5 & - & - & - & - & - & 50.1 & 25.1 & - & 37.5 & 20.6 & - & 17.6 & - & - & 23.3 & 51.3 & - & 16.3 & - & 21.2 & - & - & 17.9 & - & - & - & - & - & 28.2 \\
11 & Domain\_adaptation & - & 47.2 & 40.2 & - & 6.4 & 32.2 & 46.2 & 56.1 & 50.1 & - & - & - & - & - & - & - & - & 25.2 & - & 51.9 & 54.7 & - & - & - & - & - & - & 23.5 & - & - & - & - & 39.4 \\
12 & Zero\_Shot & - & 45.9 & 43.6 & 30.3 & - & - & - & - & 49.9 & - & - & - & - & - & - & 39.1 & 37.5 & - & 20.2 & - & - & 36.5 & 29.6 & 18.4 & - & - & - & - & - & - & - & - & 35.1 \\
13 & GT-NLP & 53.8 & - & 54.3 & 53.5 & - & 68.7 & - & - & - & 78.3 & - & 58.3 & - & - & - & - & - & - & - & - & 80.6 & - & - & - & 57.5 & - & - & 55.6 & - & - & - & - & 62.3 \\
14 & UT-NLP & - & - & - & - & 33.7 & - & 56.0 & - & 39.3 & - & - & 50.5 & - & - & - & - & - & - & - & - & - & - & - & - & - & - & - & 23.7 & - & - & - & 18.9 & 37.0 \\
15 & UncleLM & - & - & - & - & - & - & - & 60.9 & - & 70.9 & - & - & - & - & 78.3 & - & - & - & - & - & 67.0 & - & - & - & - & 36.8 & - & 55.9 & - & - & - & - & 61.6 \\
16 & Team UBD & - & - & - & - & - & - & - & - & - & 60.1 & - & 41.0 & - & - & - & - & - & 40.5 & - & 50.9 & 53.4 & - & - & - & - & - & - & 38.6 & - & - & - & - & 47.4 \\
17 & Tue-JMS & 31.3 & - & - & - & - & - & - & - & - & - & - & 50.8 & 34.7 & - & - & - & - & - & - & - & - & - & - & - & - & - & - & - & - & 10.8 & - & 13.1 & 28.1 \\
18 & UIMP-Aaman & - & - & - & - & - & 59.4 & 66.1 & 69.4 & - & - & - & - & - & - & - & - & - & 49.4 & - & - & - & - & - & - & - & - & - & - & - & - & - & - & 61.1 \\
19 & Heimerdinger & - & - & - & - & - & - & - & - & - & - & - & 60.9 & \cellcolor{blue!30}43.9 & - & - & - & - & - & - & - & - & - & - & - & - & - & - & - & - & 26.0 & - & 22.6 & 38.4 \\
20 & Lazarus NLP & - & - & - & - & - & - & - & - & - & - & - & 64.1 & 43.8 & - & - & - & - & - & - & - & - & - & - & - & - & - & - & - & - & - & - & - & 54.0 \\
21 & GOLDX & - & - & - & - & - & - & - & - & - & - & - & 42.9 & - & - & - & - & - & - & - & - & - & - & - & - & - & - & - & - & - & - & - & - & 42.9 \\
        \end{longtable}
\end{landscape}

\small
\begin{longtable}{|p{1cm}|p{3.4cm}|p{6.7cm}|p{3.2cm}|}
    \caption{Participants Information} \label{tab:participants_info} \\
    \hline
    \textbf{Tasks} & \textbf{Team} & \textbf{Affiliation} & \textbf{Paper} \\
    \hline
    \endfirsthead

    \multicolumn{4}{l}{\normalsize\textbf{Table \thetable .} Continued from previous page} \\
    \hline
    \extrasmall
    \textbf{Tasks} & \textbf{Team} & \textbf{Affiliation} & \textbf{Paper} \\
    \hline
    \endhead

    \hline
    \multicolumn{4}{l}{\textbf{Continued on next page}} \\
    \hline
    \endfoot

    \hline
    \endlastfoot

A, B & AGHNA & Institut Teknologi Bandung & \cite{abyan-etal-2025-aghna} \\ \hline
A, B, C & AKCIT & CEIA NLP & \cite{brito-etal-2025-akcit} \\ \hline
A & Amado & none & \cite{bade-etal-2025-amado} \\ \hline
A & Angeliki Linardatou & none & \cite{linardatou-etal-2025-angeliki-linardatou} \\ \hline
A & AtlasIA & AtlasIA & \cite{majjodi-etal-2025-atlasia} \\ \hline
A, B & CIC-NLP & none & \cite{abiola-etal-2025-cic-ipn} \\ \hline
A, B, C & CIOL & Shahjalal University of Science Technology, Bangladesh & \cite{hoque-etal-2025-ciol} \\ \hline
A, B & CSECU-Learners & University of Chittagong & \cite{ahmad-etal-2025-csecu-learners} \\ \hline
A, B & CSIRO-LT & CSIRO, Macquarie University & \cite{chen-etal-2025-csiro-lt} \\ \hline
A, B & CharsiuRice & Eberhard Karls University of Tübingen & \cite{yip-etal-2025-charsiurice} \\ \hline
A, B & Chinchunmei & Newcastle University, Shumei AI Research Institute & \cite{li-etal-2025-chinchunmei} \\ \hline
A & CodeBlAIzers & Queen Mary University of London & \cite{wang-etal-2025-codeblaizers} \\ \hline
A & DUT\_IR & Dalian University of Technology, China & \cite{liu-etal-2025-dut_ir} \\ \hline
A & DataBees & SSN College Of Engineering & \cite{anand-etal-2025-databees} \\ \hline
A, B, C & Deep & none & \cite{shenpo-etal-2025-deep} \\ \hline
A, B & Domain\_adaptation & Moscow Institute of Physics and Technology, University of Leeds & \cite{lepekhin-etal-2025-domain_adaptation} \\ \hline
A & EMO-NLP & Yunnan University & \cite{li-etal-2025-emo-nlp} \\ \hline
A & Emotion Train & Mohamed bin Zayed University of Artificial Intelligence & \cite{demidova-etal-2025-emotion-train} \\ \hline
A & Empaths & HSE University, Independent researcher & \cite{morozov-etal-2025-empaths} \\ \hline
A & Exploration Lab IITK & IIT Kanpur & \cite{nadeem-etal-2025-exploration-lab-iitk} \\ \hline
A & FiRC-NLP & none & \cite{tufa-etal-2025-firc-nlp} \\ \hline
A, C & GOLDX & none & \cite{clinton-etal-2025-goldx} \\ \hline
A, B, C & GT-NLP & Georgia Institute of Technology, Johnson \& Johnson, Novastraum LLC, C3 AI & \cite{saeedi-etal-2025-gt-nlp} \\ \hline
A, B & GinGer & Sharif University of Technology, Iran University of Science and Technology & \cite{naebzadeh-etal-2025-ginger} \\ \hline
A & HTU & AlHussein Technical University & \cite{saleh-etal-2025-htu} \\ \hline
A, B, C & Habib University & Habib University, Karachi, Pakistan & \cite{waheed-etal-2025-habib-university} \\ \hline
A & HausaNLP & University of the Witwatersrand Johannesburg, Federal Polytechnic, Daura & \cite{sani-etal-2025-hausanlp} \\ \hline
A, B, C & Heimerdinger & Huazhong University of Science and Technology & \cite{tong-etal-2025-heimerdinger} \\ \hline
A, C & Howard University-AI4PC & Howard University & \cite{ince-etal-2025-howard-university-ai4pc} \\ \hline
A, B, C & IASBS & IASBS University & \cite{tareh-etal-2025-iasbs} \\ \hline
A & IEGPS-CSIC & Instituto de Estudios Gallegos Padre Sarmiento (IEGPS-CSIC) & \cite{sarymsakova-etal-2025-iegps-csic} \\ \hline
A & INFOTEC-NLP & INFOTEC Centro de Investigación e Innovación en, Tecnologías de la Información y Comunicación, Secretaría de Ciencia, Humanidades, Tecnología e Innovación (SECIHTI) & \cite{santos-rodriguez-etal-2025-infotec-nlp} \\ \hline
A & ITF-NLP & Fraunhofer FKIE & \cite{kent-etal-2025-itf-nlp} \\ \hline
A, B & JNLP & National Institute of Informatics, Japan Advanced Institute of Science and Technology, National Institute of Advanced Industrial Science and Technology & \cite{xue-etal-2025-jnlp} \\ \hline
A & JUNLP\_Sarika & Jadavpur University, Kolkata, India & \cite{khatun-etal-2025-junlp_sarika} \\ \hline
A & JellyK & University of Information Technology VNU-HCM & \cite{le-etal-2025-jellyk} \\ \hline
A & KDBERT MLDistill & none & \cite{wang-etal-2025-kdbert-mldistill} \\ \hline
A, B & LATE-GIL-NLP & Universidad Nacional Autonoma de Mexico & \cite{vázquez-osorio-etal-2025-late-gil-nlp} \\ \hline
A, C & Lazarus NLP & University of New South Wales, Lazarus NLP & \cite{wongso-etal-2025-lazarus-nlp} \\ \hline
A & Lotus & PersianGulf University & \cite{ranjbar-etal-2025-lotus} \\ \hline
A & MRS & Michigan State University & \cite{afshari-etal-2025-mrs} \\ \hline
A & McGill-NLP & Université de Montréal, Mila - Quebec AI Institute, McGill University, Canada CIFAR AI Chair & \cite{verma-etal-2025-mcgill-nlp} \\ \hline
A & NCL-NLP & Newcastle University, Newcastle Upon Tyne, England & \cite{lu-etal-2025-ncl-nlp} \\ \hline
A & NITK-VITAL & National Institute of Technology Karnataka, Surathkal, India & \cite{kesanam-etal-2025-nitk-vital} \\ \hline
A & NLP-Cimat & Center for Research in Mathematics (CIMAT), Secretaría de Ciencia, Humanidades, Tecnología e Innovación (SECIHTI) & \cite{gómez-etal-2025-nlp-cimat} \\ \hline
A & NLP-DU & CSE, DU & \cite{sakib-etal-2025-nlp-du} \\ \hline
A, B & NLP\_goats & Sri Sivasubramaniya Nadar (SSN) College of Engineering & \cite{vaidyanathan-etal-2025-nlp_goats} \\ \hline
A & NTA & University of Information Technology, Vietnam National University, Ho Chi Minh City & \cite{le-etal-2025-nta} \\ \hline
A & NUST Titans & National University of Science and Technology, Islamabad Pakistan, Mid Sweden University, Sweden & \cite{khan-etal-2025-nust-titans} \\ \hline
A, B & NYCU-NLP & National Yang Ming Chiao Tung University & \cite{xu-etal-2025-nycu-nlp} \\ \hline
A & OPI-DRO-HEL & National Information Processing Institute, Warsaw, Poland & \cite{karaś-etal-2025-opi-dro-hel} \\ \hline
A, B, C & OZemi & Waseda University & \cite{takahashi-etal-2025-ozemi} \\ \hline
A, B & PAI & none & \cite{ruan-etal-2025-pai} \\ \hline
A & Pateam & PING AN LIFE INSURANCE COMPANY OF CHINA, LTD. & \cite{chen-etal-2025-pateam} \\ \hline
A, B & Pixel Phantoms & SSN college of engineering & \cite{s-etal-2025-pixel-phantoms} \\ \hline
A & PromotionGo & Hubei University & \cite{huang-etal-2025-promotiongo} \\ \hline
A & QiMP & Technical University of Munich & \cite{bogatyreva-etal-2025-qimp} \\ \hline
A & RSSN & Sri Sivasubramaniya Nadar (SSN) College of Engineering & \cite{v-etal-2025-rssn} \\ \hline
A & STFXNLP & none & \cite{murrant-etal-2025-stfxnlp} \\ \hline
B & Stanford MLab & Stanford University & \cite{le-etal-2025-stanford-mlab} \\ \hline
A, B & SyntaxMind & none & \cite{riad-etal-2025-syntaxmind} \\ \hline
A, B & TILeN & none & \cite{reyes-magaña-etal-2025-tilen} \\ \hline
A & Tarbiat Modares & Sharif University of Technology, Tarbiat Modares University & \cite{bourbour-etal-2025-tarbiat-modares} \\ \hline
A & Team A & HKBK College of Engineering, Bangalore, India, National Institute of Technology Agartala, Tripura, India & \cite{sahil-etal-2025-team-a} \\ \hline
A & Team KiAmSo & University of Tübingen & \cite{sharp-etal-2025-team-kiamso} \\ \hline
A, B, C & Team UBD & University of Bucharest & \cite{paduraru-etal-2025-team-ubd} \\ \hline
A & Team Unibuc - NLP & University of Bucharest & \cite{creanga-etal-2025-team-unibuc---nlp} \\ \hline
A & TechSSN3 & SSN College of Engineering & \cite{s-etal-2025-techssn3} \\ \hline
A, B & TeleAI & Institute of Artificial Intelligence (TeleAI), China Telecom Corp Ltd & \cite{wang-etal-2025-teleai} \\ \hline
A & Tewodros & none & \cite{bizuneh-etal-2025-tewodros} \\ \hline
B & Tewodros & none & \cite{bizuneh-etal-2025-tewodros} \\ \hline
A & Trans-Sent & Aliah University, Kolkata, Jadavpur University, Kolkata & \cite{sarif-etal-2025-trans-sent} \\ \hline
A, B, C & Tue-JMS & Universität Tübingen & \cite{han-etal-2025-tue-jms} \\ \hline
A, B, C & UB\_Tel-U & Brawijaya University, Telkom University & \cite{fatyanosa-etal-2025-ub_tel-u} \\ \hline
A, B, C & UIMP-Aaman & none & \cite{aman-parveen-etal-2025-uimp-aaman} \\ \hline
B, C & UT-NLP & none & \cite{safdarian-etal-2025-ut-nlp} \\ \hline
A, B, C & UncleLM & Iran University of Science and Technology & \cite{barfi-etal-2025-unclelm} \\ \hline
A & University of Indonesia & Universitas Indonesia' & \cite{hanif-etal-2025-university-of-indonesia} \\ \hline
A, C & UoB-NLP & University of Birmingham & \cite{leon-etal-2025-uob-nlp} \\ \hline
A & VerbaNexAI & Universidad Tecnológica de Bolívar & \cite{almanza-etal-2025-verbanexai} \\ \hline
A & XLM-Muriel & Amirkabir University of Technology, Tehran, Iran & \cite{hosseinzadeh-etal-2025-xlm-muriel} \\ \hline
A & YNU-NPCC & none & \cite{yang-etal-2025-ynu-npcc} \\ \hline
A, B, C & YNWA\_PZ & Iran University of Science and Technology & \cite{poulaei-etal-2025-ynwa_pz} \\ \hline
A, B & Zero & IIIT Hyderabad & \cite{gundam-etal-2025-zero} \\ \hline
A & Zero & IIIT-Hyderabad & \cite{gundam-etal-2025-zero} \\ \hline
A, B, C & Zero\_Shot & Chittagong University of Engineering and Technology & \cite{paran-etal-2025-zero_shot} \\ \hline
A, B & indiDataMiner & Indian Institute of Technology Guwahati, India, NTU Singapore & \cite{kumar-etal-2025-indidataminer} \\ \hline
A, B, C & maomao & none & \cite{chen-etal-2025-maomao} \\ \hline
B & \textcolor{red}{VebanexAI} & -- & \cite{moreno-etal-2025-vebanexai} \\ \hline
B & tinaal & Newcastle University, UK & \cite{zhu-etal-2025-tinaal} \\ \hline
A & wangkongqiang & none & \cite{kongqiang-etal-2025-wangkongqiang} \\ \hline
A & xiacui & Manchester Metropolitan University & \cite{cui-etal-2025-xiacui} \\ \hline
A & zhouyijiang1 & University of Yunnan & \cite{jiang-etal-2025-zhouyijiang1} \\ \hline

\end{longtable}

\end{document}